\documentclass[10pt,twocolumn,letterpaper]{article}

\usepackage{iccv}
\usepackage{times}
\usepackage{epsfig}
\usepackage{graphicx}
\usepackage{amsmath}
\usepackage{amssymb}
\usepackage{color}

\usepackage{multirow}
\usepackage{booktabs}

\makeatletter
\def\ps@myheadings{%
    \let\@oddfoot\@empty\let\@evenfoot\@empty
    \def\@evenhead{\thepage\hfil\slshape\leftmark}%
    \def\@oddhead{{\slshape\rightmark}\hfil\thepage}%
    \let\@mkboth\@gobbletwo
    \let\sectionmark\@gobble
    \let\subsectionmark\@gobble
    }
  \if@titlepage
  \renewcommand\maketitle{\begin{titlepage}%
  \let\footnotesize\small
  \let\footnoterule\relax
  \let \footnote \thanks
  \null\vfil
  \vskip 60\p@
  \begin{center}%
    {\LARGE \@title \par}%
    \vskip 3em%
    {\large
     \lineskip .75em%
      \begin{tabular}[t]{c}%
        \@author
      \end{tabular}\par}%
      \vskip 1.5em%
    {\large \@date \par}%       % Set date in \large size.
  \end{center}\par
  \@thanks
  \vfil\null
  \end{titlepage}%
  \setcounter{footnote}{0}%
}
\else
\renewcommand\maketitle{\par
  \begingroup
    \renewcommand\thefootnote{\@fnsymbol\c@footnote}%
    \def\@makefnmark{\rlap{\@textsuperscript{\normalfont\@thefnmark}}}%
    \long\def\@makefntext##1{\parindent 1em\noindent
            \hb@xt@1.8em{%
                \hss\@textsuperscript{\normalfont\@thefnmark}}##1}%
    \if@twocolumn
      \ifnum \col@number=\@ne
        \@maketitle
      \else
        \twocolumn[\@maketitle]%
      \fi
    \else
      \newpage
      \global\@topnum\z@   % Prevents figures from going at top of page.
      \@maketitle
    \fi
    \thispagestyle{plain}\@thanks
  \endgroup
  \setcounter{footnote}{0}%
}
\makeatother

\usepackage[breaklinks=true,bookmarks=false]{hyperref}

\iccvfinalcopy % *** Uncomment this line for the final submission

\def\iccvPaperID{5344} % *** Enter the ICCV Paper ID here

\begin{document}

%%%%%%%%% TITLE
\title{On the Efficacy of Knowledge Distillation}

\author{Jang Hyun Cho\\
Cornell University\\
{\tt\small jc2926@cornell.edu}
\and
Bharath Hariharan\\
Cornell University\\
{\tt\small bh497@cornell.edu}
}

\maketitle
\ificcvfinal\thispagestyle{empty}\fi

\begin{abstract}
   In this paper, we present a thorough evaluation of the efficacy of knowledge distillation and its dependence on student and teacher architectures. Starting with the observation that more accurate teachers often don't make good teachers, we attempt to tease apart the factors that affect knowledge distillation performance. We find crucially that larger models do not often make better teachers. We show that this is a consequence of mismatched capacity, and that small students are unable to mimic large teachers. We find typical ways of circumventing this (such as performing a sequence of knowledge distillation steps) to be ineffective. Finally, we show that this effect can be mitigated by stopping the teacher's training early. Our results generalize across datasets and models.  
%\textcolor{red}{Despite the rapidly growing interest on knowledge distillation, factors that affect the performance of distilled student have rarely been studied. In this paper, we present a thorough evaluation of the efficacy of knowledge distillation and its dependence on the capacity of the student network.  }

\end{abstract}
\begin{figure}
\centering
\includegraphics[width=\linewidth]{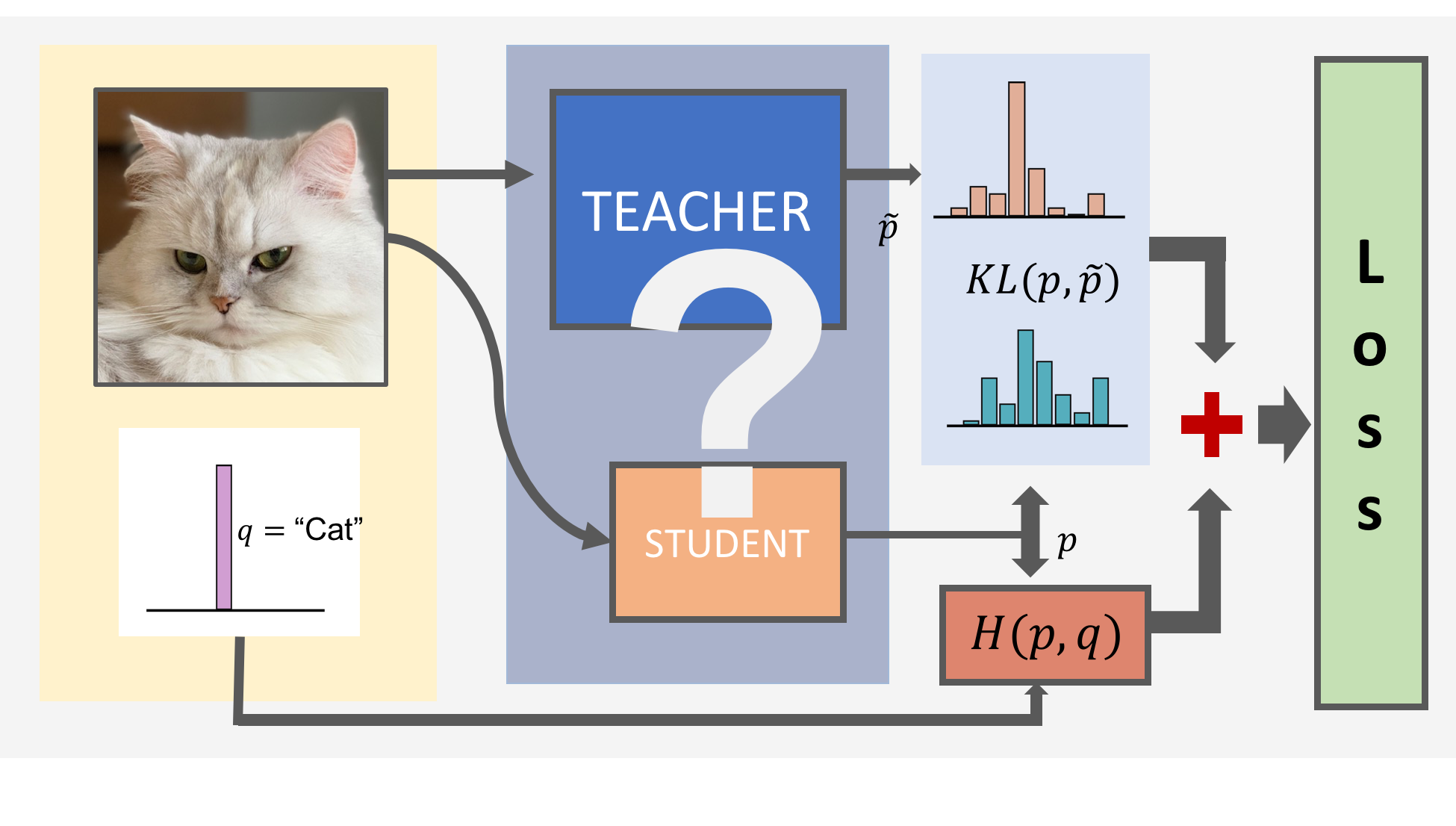}
\caption{An illustration of standard knowledge distillation. Despite widespread use, an understanding of when the student can learn from the teacher is missing. }
\label{fig:cat}
\end{figure}

%\todo{1) Finish abstract. 2) Make sure to include enough results with depth factor so it does not seem like we only consider width. 3) Emphasize "bias": teacher's logit (KD loss) acts as learning bias if student is not capable of optimize well. 4) Fix texts inside figures and add some cooler figures/ illustration. }

%%%%%%%%% BODY TEXT
\section{Introduction}
The past few years have seen dramatic improvements in visual recognition systems, but these improvements have been driven by deeper and larger convolutional network architectures.
The large computational complexity of these architectures has limited their use in many downstream applications. As such, there has been a lot of recent research on achieving the same or similar accuracy with smaller models.
Some of this work has involved building more efficient neural network families~\cite{densenet, morphnet}, pruning weights from larger neural networks~\cite{han2015deep}, quantizing existing networks to use fewer bits for weights and activations~\cite{wu2016quantized} and distilling knowledge from larger networks into smaller ones~\cite{KD,ba2014deep}. 

The last of these, knowledge distillation, is a general-purpose technique that at first glance is widely applicable and complements all other ways of compressing neural networks~\cite{mishra2018apprentice}. 
The key idea is to use soft probabilities (or `logits') of a larger ``teacher network'' to supervise a smaller ``student'' network, in addition to the available class labels.
These soft probabilities reveal more information than the class labels alone, and can purportedly help the student network learn better.

The appeal of this approach is in its apparent generality: \emph{any} student can learn from \emph{any} teacher.
But does knowledge distillation fulfill this promise of generality?
Unfortunately, in spite of the recent interest in variants of knowledge distillation~\cite{lan2018knowledge,YimCVPR2017,yang2018knowledge, tarvainen2017mean, Zhang_2018_CVPR,attentiontransfer,NIPS2017_6676, NST}, an empirical answer to this question is missing. 
Prior experiments have typically looked at a small number of carefully chosen architectures, with the implicit assumption that conclusions will generalize across student or teacher architectures.
However, there are a few isolated reports of failed experiments with knowledge distillation that suggest that this might not be true.
For example, Zagoruyko and Komodakis observe that they are ``unable to achieve positive results with knowledge distillation on ImageNet''~\cite{attentiontransfer}.
What characterizes this, and other experiments where knowledge distillation does not seem to improve performance?
Are there student-teacher combinations that perform better?
And finally, is there something we can do to improve performance for other combinations?

In this paper, we seek to answer these questions.
We find that in general, the teacher accuracy is a poor predictor of the student's performance. Larger teachers, though they are more accurate by themselves, do not necessarily make for better teachers.
We explore the reasons for this and demonstrate that as the teacher grows in capacity and accuracy, the student often finds it difficult to emulate the teacher (resulting in a high KL divergence from the teacher logits even during training).
We show that this issue cannot be mitigated by solutions suggested in prior work, such as using a sequence of  knowledge-distillation steps to increase the student accuracy.
Finally, we find an effective solution to the problem: regularizing the teacher by stopping the training of the teacher early, and stopping knowledge distillation close to convergence to allow the student to fit the training loss better.
Our solution is simple to implement and effective across the board at improving the efficacy of knowledge distillation.

The rest of the paper is organized as follows.
After describing related work (Sec.~\ref{sec:relwork}), we first provide some background on knowledge distillation and attention transfer (Sec.~\ref{sec:background}).
We then describe our experimental setting(Sec.~\ref{sec:methods}).
In Sec.~\ref{sec:results} we discuss our findings and the empirical evidence for each.

\section{Related Work}
\label{sec:relwork}
\subsection{Knowledge distillation}
The notion of training smaller, cheaper models (``students'') to mimic larger ones (``teachers'') is an old one, first described in a seminal paper on model compression by Bucilu\v{a} et al~\cite{modelcompression}.
This technique can be applied to deep neural networks almost out of the box~\cite{ba2014deep, KD}.
In this paper, we use the knowledge distillation framework described by Hinton et al.~\cite{KD}.
A brief description of knowledge distillation is provided in Section~\ref{sec:background}.
The original paper on knowledge distillation experimented with the idea on a few small datasets, but a thorough empirical evaluation of knowledge distillation is missing.
% The original paper on knowledge distillation experimented with the idea on a few small datasets, but a thorough empirical evaluation of knowledge distillation and its dependence on student or teacher architectures is missing.

Meanwhile, the focus of past work has been either on improving the quality of knowledge distillation or finding new applications for the idea.
On the former direction, prior work has explored adding additional losses on intermediate feature maps of the student to bring them closer to those of the teacher~\cite{attentiontransfer,NS,YimCVPR2017}.
Zhang et al. train a pair of models, distilling knowledge bidirectionally at every epoch ~\cite{Zhang_2018_CVPR}. 
Tarvainen et al. find that averaging consecutive student models over training steps tend to produce better performing students~\cite{tarvainen2017mean}.
Yang et al. modify the loss function of teacher network to be more ``tolerant'' (that is, by adding more terms to make the model intentionally maintain high \textit{energy}, benefiting from teacher's misclassified logits)~\cite{yang2018knowledge}.

% \textcolor{blue}{\textbf{COMMENT} Maybe merge the next two paragraphs and discuss about bias-stacking issue? (when student capacity is limited and learning from teacher is too challenging, KD acts as a form of bias and it stacks up with generations).}

A particular approach to improving knowledge distillation is to perform knowledge distillation repetitively (we call it \emph{sequential knowledge distillation}~\cite{yang2018knowledge,furlanello2018born,lan2018knowledge}). 
A particular way of using sequential knowledge distillation is as an alternative to ensembling to increase model accuracy~\cite{lan2018knowledge,furlanello2018born}.
For example, Furlanello et al.~\cite{furlanello2018born} suggest training an ensemble of networks using a sequence of knowledge-distillation steps where a network uses its own previous version as a teacher.
Interestingly, our results suggest that this approach \emph{underperforms} an ensemble trained from scratch, and furthermore, such sequential knowledge distillation reduces the ability of the network to act as a teacher.
More generally, we find that these methods are highly dependent on the student capacity. 
In fact we find them ineffective in many cases particularly when student capacity is limited or the dataset is complex.
% However, this method is highly domain-dependent and in fact we find it ineffective in more complex dataset such as ImageNet~\cite{imagenet}.

In terms of applications of knowledge distillation, prior work has found knowledge distillation to be useful for sequence modeling~\cite{sequence, speech}, semi-supervised learning~\cite{tarvainen2017mean}, domain adaptation~\cite{Meng2018adversarial}, multi-modal learning~\cite{gupta2016cross} and so on.
This wide applicability of the idea of knowledge distillation makes an exhaustive evaluation of knowledge distillation ideas even more important.

% Adding for ICCV starts here -----
% Talk about papers that use KD in their paper and their applications. (Medical, NLP, etc.). (TODO)
% Adding for ICCV ends here -----

% TODO: Address more papers regarding sequential KD

%------------------------------------------------------------------------ 

% \subsection{Incremental learning}
% Recently, Incremental learning has been proposed as a variation of knowledge distillation~\cite{Zhang_2018_CVPR,furlanello2018born,sequence,mirzadeh2019improved}. Here the main objective is not the network compression as knowledge distillation is originally designed for~\cite{KD}, but the performance improvement given the same network and dataset. 

\section{Background: Knowledge distillation}
\label{sec:background}
% There have been many variants of knowledge distillation proposed in the literature~\cite{NS, attentiontransfer, KD, ba2014deep}.
% In this paper, we consider two variants of knowledge distillation: vanilla knowledge distillation~\cite{KD}, and attention transfer~\cite{attentiontransfer}.
% For completeness, we briefly review both knowledge distillation and attention transfer below.

The key idea behind knowledge distillation is that the soft probabilities output by a trained ``teacher'' network contains a lot more information about a data point than just the class label.
For example, if multiple classes are assigned high probabilities for an image, then that might mean that the image must lie close to a decision boundary between those classes.
Forcing a student to mimic these probabilities should thus cause the student network to imbibe some of this knowledge that the teacher has discovered above and beyond the information in the training labels alone.

Concretely, given any input image $x$ the teacher network produces a vector of scores $\mathbf{s}^t(x) = [s^t_1(x), s^t_2(x), \ldots, s^t_K(x)]$
that are converted into probabilities: $
p^t_k(x) = \frac{e^{s^t_k(x)}}{\sum_j e^{s^t_j(x)}}
$.
Trained neural networks produce peaky probability distributions, which may be less informative.
%For trained neural networks, these probabilities can be quite peaky, bordering on one-hot encodings of the class label.
%As such, they may not be very informative as is for the student to learn from.
Hinton et al~\cite{KD} therefore propose to ``soften'' these probabilities using temperature scaling~\cite{guo2017calibration}:
\begin{align}
\tilde{p}^t_k(x) = \frac{e^{s^t_k(x)/\tau}}{\sum_j e^{s^t_j(x)/\tau}}
\end{align}
where $\tau>1$ is a hyperparameter.

A student similarly produces a softened class probability distribution, $\tilde{\mathbf{p}}^s(x)$.
The loss for the student is then a linear combination of the typical cross entropy loss $\mathcal{L}_{cls}$ and a knowledge distillation loss$\mathcal{L}_{KD}$:
\begin{align*}
\mathcal{L} &= \alpha \mathcal{L}_{cls} + (1-\alpha)\mathcal{L}_{KD}\\
    \text{where}\quad \mathcal{L}_{KD} &= - \tau^2\sum_k \tilde{p}_k^t(x) \log \tilde{p}_k^s(x)
\end{align*}
$\alpha$ and $\tau$ are hyperparameters; popular choices are $\tau \in \{3, 4, 5\}$ and $\alpha = 0.9$~\cite{attentiontransfer,NST,lan2018knowledge, KD}.

\section{Methods}
\label{sec:methods}
We perform experiments on both CIFAR10 and ImageNet.
In each case we keep the student the same and use multiple teachers of varying capacity to perform knowledge distillation.

\paragraph{CIFAR10}
For experiments on CIFAR10, we run each model for 200 epochs using SGD with momentum $0.9$ and set the initial learning rate $\gamma=0.1$, dropping $0.2$ every 60 epochs. Standard data augmentation was applied to the dataset. For the hyperparameters regarding knowledge distillation, we stayed consistent with the popular choice (\cite{KD}, \cite{attentiontransfer}): Temperature $\tau=4$, $\alpha=0.9$, and $\beta=1000$ for attention transfer. The same experiment was repeated 5 times and median, mean, and standard deviation are reported. We consider three different network architectures: ResNet~\cite{resnet}, WideResNet~\cite{wideresnet}, and DenseNet~\cite{densenet}. 

\paragraph{ImageNet}
For ImageNet experiments we followed Zagoruyko et al.~\cite{attentiontransfer} closely since it was the first successful work of knowledge distillation on ImageNet, to the best of our knowledge. We used SGD with nesterov momentum $0.9$, initial learning rate $\gamma=0.1$, weight decay $1\times 10^{-4}$ , and dropped learning rate by $0.1$ every 30 epochs. As with CIFAR10, we set temperature $\tau=4$, $\alpha=0.9$, and $\beta=1000$ for attention transfer. For ImageNet experiments, we consider ResNet~\cite{resnet}.% and WideResNet~\cite{wideresnet}. 

\begin{figure*}
\centering
\includegraphics[width=0.49\linewidth]{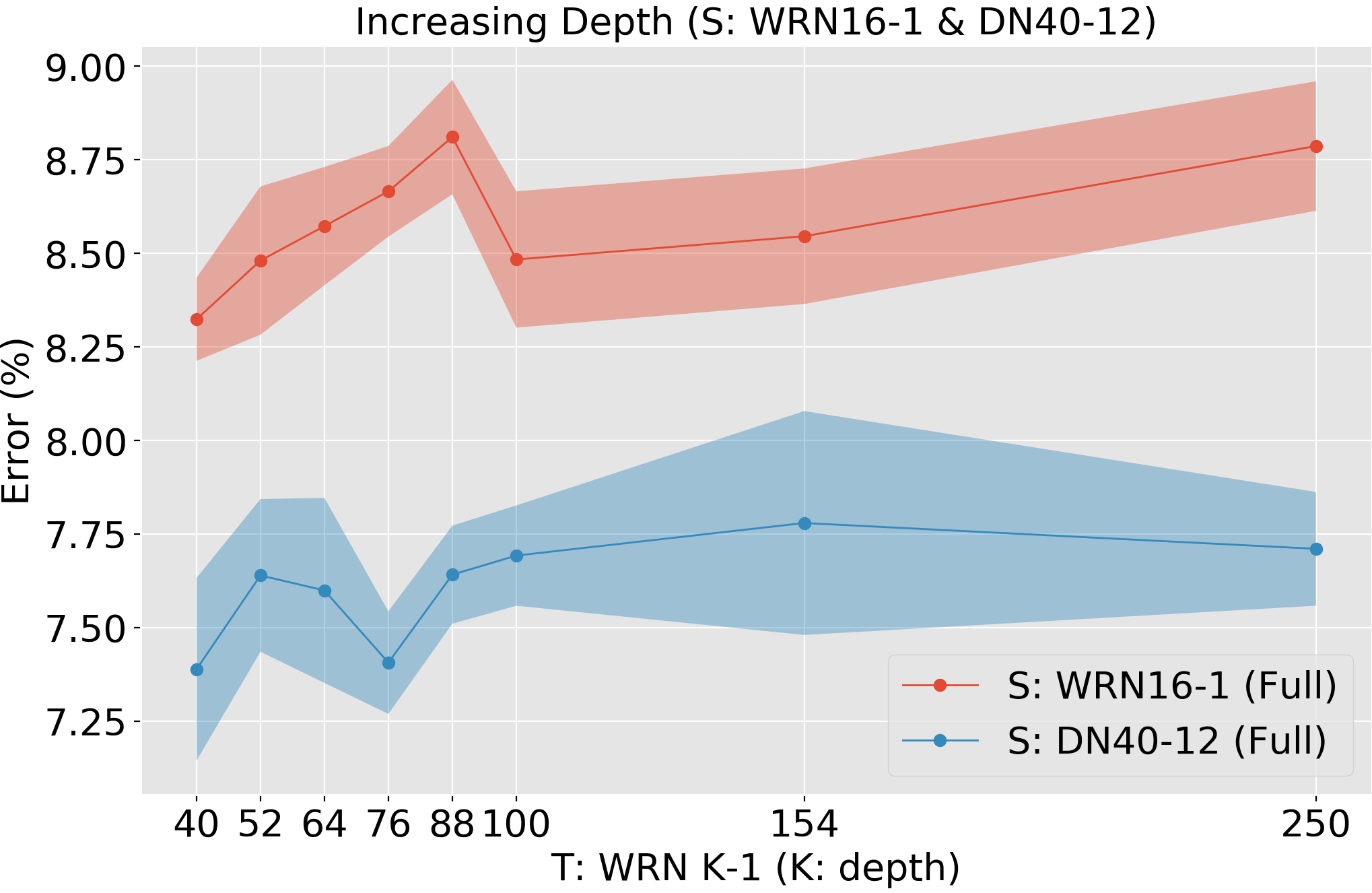}
\includegraphics[width=0.49\linewidth]{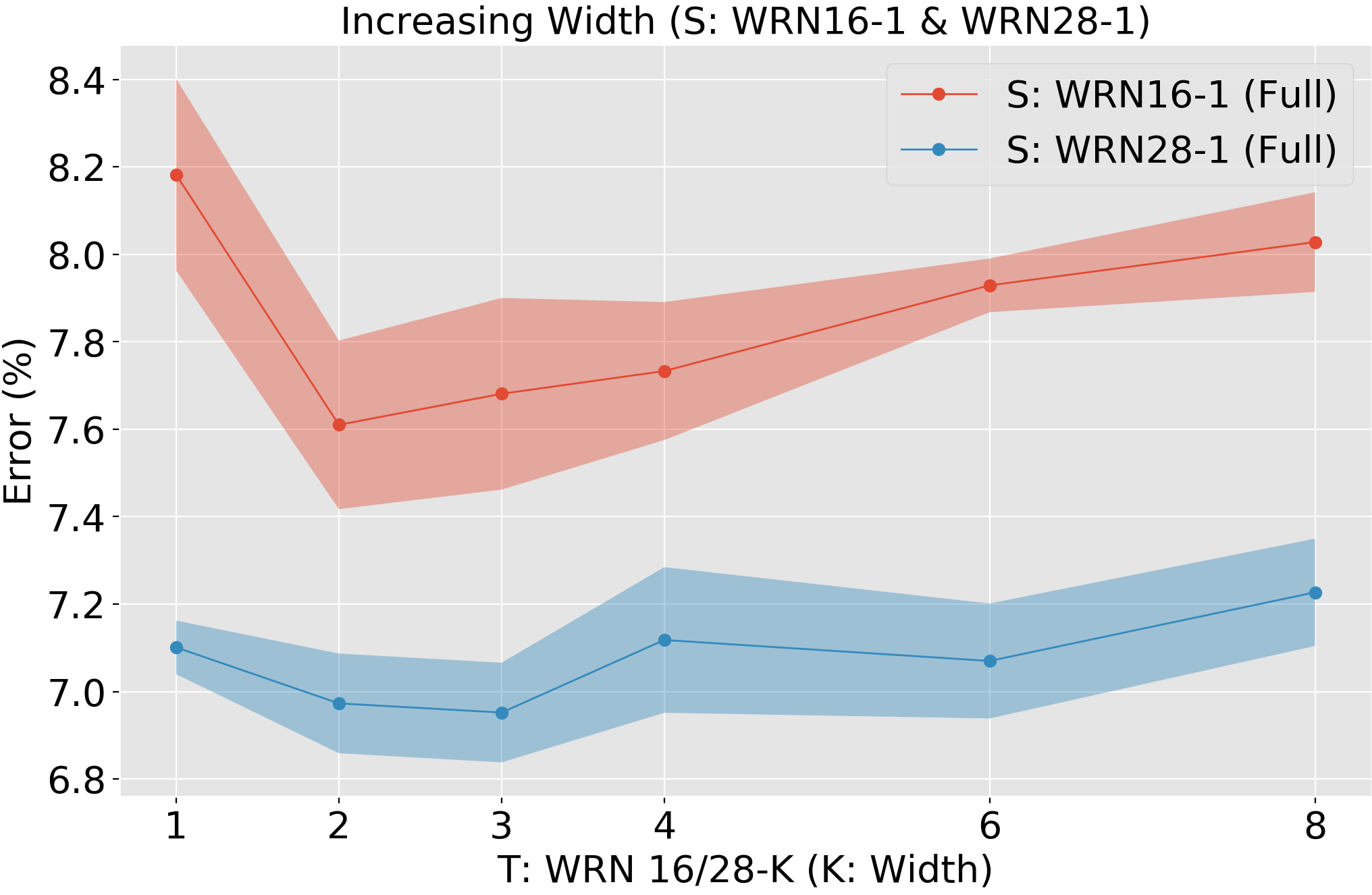}
\caption{The error plot of student networks distilled from different teachers on CIFAR10. WideResNet~\cite{wideresnet} 16-1 (\textbf{left/red}, \textbf{right/red}), 28-1 (\textbf{right/blue)}), and DenseNet~\cite{densenet} 40-12 (\textbf{left/blue}) were used as the student networks. Increasing teacher capacity  (depth: \textbf{left}, width: \textbf{right}) and thus accuracy does not necessarily increase the accuracy of the student network, indicating that the accuracy of the teacher network alone is not a valid metric to knowledge distillation. }
\label{fig:TSoptimization}
\end{figure*}
\section{Results}
\label{sec:results}
\subsection{Bigger models are not better teachers}

The idea behind knowledge distillation is that soft probabilities from a trained teacher reflect more about the data than the true label alone. One might expect that as the teacher becomes more accurate, these soft probabilities will capture the underlying class distribution better and thus serve as better supervision to the student.
Thus, intuitively, we might expect that bigger, more accurate models might form better teachers.

We first evaluate if this is true on CIFAR10 dataset. In Figure~\ref{fig:TSoptimization},
The \textbf{red} and \textbf{blue} lines shows the accuracy for different student networks trained from different teachers; the left plot varies the ``depth'' of the teacher while the right plot varies the ``width''.
% We first evaluate if this is true on CIFAR10.
% Table~\ref{tab:bigger} shows the results for two variants of knowledge distillation (with and without attention transfer), two students (WRN16-1 and WRN28-1) and multiple teachers of increasing size and accuracy.
% Figure~\ref{fig:TSoptimization} plots the accuracy for two students trained with different teachers.
From these experiments, we find that the hypothesis that bigger, more accurate models make better teachers is incorrect: although the teacher accuracy continues to rise as the teacher becomes larger (see supplementary for teacher accuracies), the student accuracy rises and then begins to fall. 
% There is a ``sweet spot'' in terms of the relative size of the teacher and student that leads to the best performance.
One might wonder if this is an artifact of the CIFAR dataset.
We repeated the experiment on ImageNet, with ResNet18 as the student and ResNet18, ResNet34, ResNet50, and ResNet152 as teachers.
The results are shown in Table~\ref{tab:biggerimnet}.
As can be seen, as the teacher becomes larger and more accurate, the student becomes \emph{less accurate}. 
% For ImageNet, there does not seem to be a sweet spot at all.

\begin{table}
\begin{center}
\resizebox{0.9\linewidth}{!}{
\begin{tabular}{lcc}
\toprule
Teacher & Teacher Error (\%) & Student Error (\%) \\
\midrule
- & - & 30.24\\
ResNet18 & 30.24 & 30.57\\
ResNet34 & 26.70 & 30.79 \\
ResNet50 & 23.85 & 30.95\\
\bottomrule
\end{tabular}
}
\end{center}
\caption{Top-1 error rate for various teachers for a ResNet18 student on ImageNet. The first row corresponds to training from scratch.}
\label{tab:biggerimnet}
\end{table}

What might be the reason for this decrease?
One possibility is that as the teacher becomes both more confident and more accurate, the output probabilities start resembling more and more a one-hot encoding of the true label, and thus the information available to the student decreases. However, softening the probabilities with high temperature~\cite{KD} did not change this result (detailed later in Figure~\ref{fig:ablation}, Table~\ref{tab:highT}), invalidating this hypothesis. 
Below, we propose an alternative hypothesis.

\subsection{Analyzing student and teacher capacity}
\begin{table}
\begin{center}
\resizebox{0.8\linewidth}{!}{
\begin{tabular}{llcc}
\toprule
Student & Teacher & KD Error & KD Error \\
& & (\%,Train) & (\%,Test) \\
\midrule
\multirow{4}{*}{WRN28-1}    & WRN28-3 & 0.23 & 4.05 \\
    & WRN28-4 & 0.25 & 4.53 \\
    & WRN28-6 & 0.23 & 4.54 \\
    & WRN28-8 & 0.31 & 4.81 \\
\midrule
\multirow{4}{*}{WRN16-1}    & WRN16-3 & 1.70 & 6.32 \\
    & WRN16-4 & 1.69 & 6.52 \\
    & WRN16-6 & 1.94 & 6.91 \\
    & WRN16-8 & 1.69 & 7.01 \\
\bottomrule
\end{tabular}
}
\end{center}
\caption{KD error on CIFAR10 for multiple teachers and students. The supplementary shows similar results from teachers with increasing depth. }
\label{tab:kderror}
\end{table}
There might be two reasons why a larger, more accurate teacher doesn't lead to better student accuracy:
\begin{enumerate}
\item The student is able to mimic the teacher, but this does not improve accuracy. This would suggest a mismatch between the KD loss and the accuracy metric we care about.
\item The student is unable to mimic the teacher, suggesting a  mismatch between student and teacher \emph{capacities}.
\end{enumerate}

We evaluated these hypotheses on CIFAR10 and ImageNet.
In Table~\ref{tab:kderror}, we show the \emph{KD error} for CIFAR: the fraction of examples for which the student and teacher predictions differ.
Odd rows in Table~\ref{tab:eskd} show the KD loss on ImageNet for a ResNet 18 student trained with different teachers.
(We show KD error instead of KD loss on CIFAR because of scale issues caused by peaky output distributions).

In both cases, the KD Error/Loss is much higher for the largest teacher, which in turn leads to the least accurate student.
This suggests that the student \emph{is unable to mimic large teachers} and points to the second hypothesis, namely, the issue is one of mismatched capacity. 
We therefore posit that 
on both ImageNet and CIFAR, due to much lower capacity, the student is unable to find a solution in its space that corresponds well to the largest teacher.

% We observe a similar effect for CIFAR10 in Table~\ref{tab:kderror}. For CIFAR, the KD loss itself was a poor indicator of the student's ability to emulate the teacher because of variations in the teachers' confidence.
% We therefore looked at the \emph{KD error}: the fraction of examples for which the student and teacher predictions differ.
% The KD error for both train and test for two different students and multiple teachers is shown in Table~\ref{tab:kderror}.
% The KD error on the test set increases with teacher size, indicating that the student is indeed unable to match the teacher.
% Interestingly, the KD error stays low on train set, indicating that the student might mimic the teacher during training but struggles during testing. \textcolor{red}{This overfitting to the particular teacher's logits would act as a bias in \textit{sequential knowledge distillation} which we will discuss more in the next section.}
% In any case, the key observation is that on both CIFAR10 and ImageNet, small students struggle to mimic large teachers.
% \textcolor{red}{The issue is therefore the limited model capacity of the student network for both CIFAR and ImageNet experiments.}

\subsection{Distillation adversely affects training}
Note that knowledge distillation performs particularly poorly on ImageNet, where \emph{all teachers} lead to lower student accuracy than a student trained from scratch (Table~\ref{tab:biggerimnet}).
While the previous section suggests that the student may not have enough capacity to match a very large teacher, it is still a mystery why \emph{no teacher} improves accuracy on ImageNet.
Despite multiple recent papers in knowledge distillation, experiments on ImageNet are rarely reported.
The few that do report find that  standard setting of knowledge distillation fails on ImageNet~\cite{attentiontransfer} or perform an experiment with a small portion of ImageNet~\cite{tarvainen2017mean}.
But the reason for this has not been explored. 

\begin{figure}
\centering
\includegraphics[width=\linewidth]{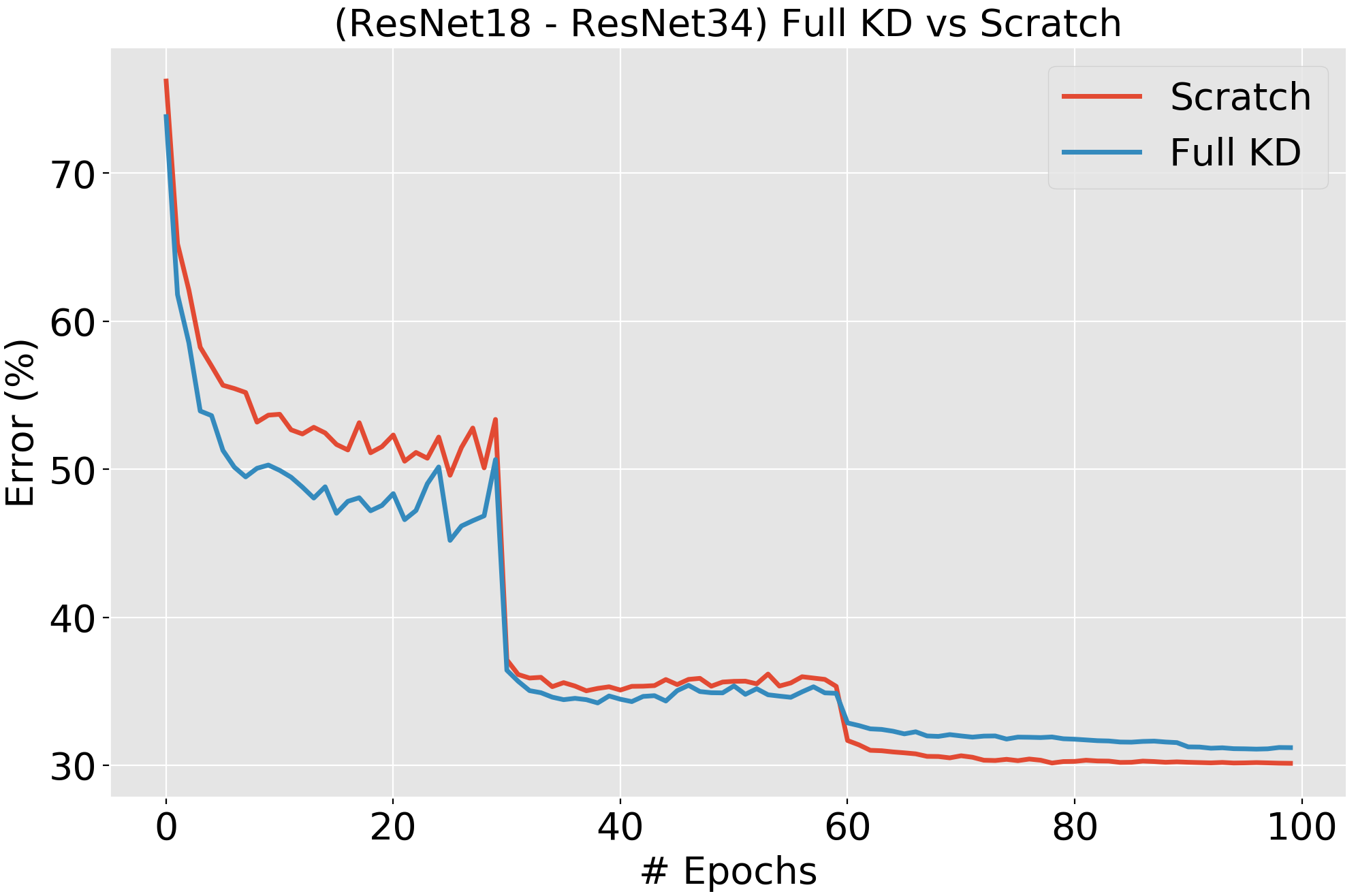}
\caption{Imagenet result of error plot of full knowledge distillation and training from scratch. In the figure the student is trained with ResNet34. 
% The relative error between the two decreases as the learning rate decays, 
Knowledge distillation helps initially but starts to hurt accuracy later in training. The same behavior occurs in the plots with different teachers (more plots in Supplementary). }
\label{fig:fkd_eskd}
\end{figure}

We dug deeper into the result.
Figure~\ref{fig:fkd_eskd} shows a comparison of validation accuracy plots between ResNet18 trained from scratch and using knowledge distillation with ResNet34. 
We find that while the KD loss improves validation accuracy initially, it begins to hurt accuracy towards the end of training.
% We first examined our ImageNet result, and looked for explanations in prior work.
% 

%We found that training on ImageNet introduces nuances that are not observed in CIFAR.

We hypothesized that because ImageNet is a more challenging problem, the low-capacity student may be in the underfitting regime.
The student may not have enough capacity to minimize both the training loss and the knowledge distillation loss, and might end up minimizing one loss (KD loss) at the expense of the other (cross entropy loss), especially towards the end of training.

% . Balancing two losses might cause it to lower one loss (the KD loss) at the expense of the other (the cross entropy loss) \textcolor{red}{especially near convergence when the learning rate has been decayed to low (i.e., in the phase where the network is being fine-tuned to optimize the both loses). Once student is no longer optimize both loss terms, the teacher's logits start to act as a \textit{bias} rather than a \textit{guidance}, potentially harming the final performance of the student. }
% In this situation, forcing it to both match the ground truth as well as the predictions of a teacher may be counter-productive: The student might do one better at the expense of the other.
% This might cause the classification loss (and thus the error rate) to actually \emph{increase} as the student tries to minimize the KL divergence between its predictions and the teacher. 
This hypothesis suggests that we might want to stop the knowledge distillation early in the training process, and do gradient descent only on the cross-entropy loss for the rest of the training. 
We call this process ``Early-stopped'' knowledge distillation (``ESKD'') as opposed to standard knowledge distillation (``Full KD''). 
%From now, ``ESKD'' refers to ``early-stopped knowledge distillation,'' while ``Full KD'' refers to a standard knowledge distillation~\cite{KD}.

Table~\ref{tab:eskd} shows how this version compares to standard knowledge distillation, and also shows the loss values at the end of training.
%To see if our hypothesis about the knowledge distillation loss driving the cross entropy loss higher is true, we also show both loss values at the end of training.
We find that the early-stopped version is better for all three teachers.
We also find that, consistent with our hypothesis, the early-stopped version achieves a \emph{lower} training cross-entropy loss and a \emph{higher} KD loss than the baseline version suggesting that the latter models are indeed trading off one loss against the other.
% training on both cross entropy and knowledge distillation throughout causes the model to sacrifice classification performance: it ends up with a higher cross entropy loss than its counterpart where knowledge distillation was stopped early.
% Thus, for ResNet34 and ResNet50, even though the former ends up being a better imitator of the teacher (lower knowledge distillation loss),  the latter is more accurate.
Note also that this simple trick of stopping knowledge distillation early now gives us the promised benefit of knowledge distillation: all the early-stopped students in Table~\ref{tab:eskd} perform better than a model of similar architecture trained from scratch (30.24\% accuracy).

However, early stopping does not change our original observation: that larger, more accurate teachers don't result in more accurate students.
Even with early sstopping, we find that the KD loss on the test set \emph{increases} with increasing teacher size, suggesting that the student is still struggling to mimic the teacher, and that it is indeed an issue of student capacity. 

% Note that this opposition between the KD loss and cross entropy is not the case in CIFAR or other light-weighted datasets, where the small network is capable of optimizing both CE loss and KD loss resulting in a benefit with full training with knowledge distillation. This suggests that the difficulty of the task affects the performance of knowledge distillation.

\begin{table}
\begin{center}
\resizebox{0.99\linewidth}{!}{
\begin{tabular}{lcccc}
\toprule
Teacher & Top-1 Error  & CE & KD & KD \\
& (\%, Test) & (Train) & (Train) & (Test)\\
\midrule
ResNet18 & 30.57 & 0.146 & 2.916 & 3.358 \\
ResNet18 (ES KD) & 29.01 & 0.123 & 2.234 & 2.491 \\
\midrule
ResNet34 & 30.79 & 0.145 & 1.357  & 1.503\\
ResNet34 (ES KD) & 29.16 & 0.123 & 2.359 & 2.582\\
\midrule
ResNet50 & 30.95 & 0.146 & 1.553 & 1.721\\
ResNet50 (ES KD) & 29.35 & 0.124 & 2.659 & 2.940\\
\bottomrule
\end{tabular}%
}
\end{center}
\caption{Early-stopping the knowledge distillation can prevent the student from degrading its classification performance on ImageNet. }
\label{tab:eskd}
\end{table}

\begin{table*}
\begin{center}
\resizebox{0.9\linewidth}{!}{
\begin{tabular}{lcccccc}
\toprule
Model  & \# Params & Method &Last Gen. Err. & All Gen. Ensmeble Err. & Scratch Err. & Scratch Ensemble Err.  \\
\midrule
ResNet8 &0.07M&AT+KD&13.469&12.786&12.569*& \textbf{10.176} \\
ResNet14 &0.17M&AT+KD&9.226&8.653&9.078*& \textbf{6.675} \\
WRN16-2 &0.69M&KD&6.101&5.181&6.428& \textbf{4.865} \\
WRN16-2 &0.69M&AT+KD&5.696&5.310&6.418& \textbf{5.003} \\
\bottomrule
\end{tabular}
}
\end{center}
\caption{5 generations of knowledge distillation were done, and the errors for the last generation of distillation sequence (``Last Gen. Err.''), ensemble of all generations (``All Gen. Ensemble Err.''), first generation (``Scratch Err.''), and the ensemble of the same number of scratch models (``Scratch Ensemble Err.'') were reported. Errors with (*) show the cases where repeating knowledge distillation even decreased the performance, and for all models support the claim that repeating knowledge distillation is ineffective.}
\label{tab:sequential}
\end{table*}

\begin{table}
\begin{center}
\resizebox{0.8\linewidth}{!}{
\begin{tabular}{lcc}
\toprule
Teacher & Teacher & Student \\
Training & Error (\%) & Error (\%) \\
\midrule
\multirow{2}{*}{Scratch} & \multirow{2}{*}{5.34} & 7.61 \\
& & (7.68 $\pm$ 0.259) \\
\multirow{2}{*}{5 KD iterations} & \multirow{2}{*}{4.89} & 7.79 \\
& & (7.67 $\pm$ 0.19) \\
\bottomrule
\end{tabular}
}
\end{center}
\caption{Sequential knowledge distillation does not make better teachers even when it improves accuracy. The student is WRN16-1, which achieves an error of $(8.759 \pm 0.129)$ when trained from scratch. The teacher is WRN16-3.}
\label{tab:sequentialteacher}
\end{table}

\begin{table}
\begin{center}
\resizebox{0.99\linewidth}{!}{
\begin{tabular}{lccc}
\toprule
\multirow{2}{*}{Training Procedure} & Large & Medium & Small  \\
& Error (\%) & Error (\%)  &Error (\%)\\
\midrule
\multirow{2}{*}{Large$\rightarrow$Med.$\rightarrow$ Small} & \multirow{2}{*}{4.41} & \multirow{2}{*}{4.80} & 8.04 \\
& & & $(7.99 \pm 0.24)$ \\
\multirow{2}{*}{Med. $\rightarrow$ Small} & \multirow{2}{*}{-} & \multirow{2}{*}{5.34} & \textbf{7.614} \\
& & & $(\mathbf{7.68 \pm 0.26})$ \\
\multirow{2}{*}{Large $\rightarrow$ Small} & \multirow{2}{*}{4.41} & \multirow{2}{*}{-} & 7.98 \\
& & & $(8.03 \pm 0.14)$\\
\bottomrule
\end{tabular}%
}
\end{center}
\caption{Using sequential knowledge distillation to distill from a large model (WRN16-8) to a medium model (WRN16-3), and from the latter to a small model (WRN16-1) does not help. The optimal approach still is to distill directly from the medium model to the small model, even though the teacher in this case has lower accuracy.}
\label{tab:largetosmall}
\end{table}

\begin{table}
\begin{center}
\resizebox{0.999\linewidth}{!}{
\begin{tabular}{lccc}
\toprule
Training procedure & 1st Teacher & 2nd Teacher & Student  \\
& Error (\%) & Error (\%)  &Error (\%)\\
\midrule
Large $\rightarrow$Small$\rightarrow$ Small & 21.69 & 29.45 & 29.41 \\
% & & &  \\
Med. $\rightarrow$Small$\rightarrow$ Small & 23.85 & 29.35 & 29.35 \\
% & & &  \\
Small $\rightarrow$Small$\rightarrow$ Small & 30.24 & \textbf{29.01} & 29.15\\
% & & &  \\
\midrule
Small $\rightarrow$Small $\rightarrow$ Small~\cite{yang2018knowledge}  & - & -& 30.12$^*$\\
Small $\rightarrow$Small$\times 5$~\cite{yang2018knowledge}  & - & -& 29.60$^*$\\
\bottomrule
\end{tabular}%
}
\end{center}
\caption{Imagenet experiment of sequential early-stopped knowledge distillation (ESKD). ``2nd Teacher'' is first distilled from ``1st Teacher'', and then ``Student'' is trained with the ``2nd Teacher''. The last two lines compare with other variants of sequential knowledge distillation. [*] indicates that the number is inherited from the original paper. }
\label{tab:imagenetsequential}
\end{table}

\subsection{The efficacy of repeated knowledge distillation}
\label{sec:sequential}

If the difference between teacher and student capacities is very large, one possibility is to first distill from the large teacher to an intermediate teacher and then distill to the student, so that each knowledge distillation step has a better match between student and teacher capacity.
This notion of sequential knowledge distillation has been proposed in the literature in other contexts.
Recently Furlanello et al~\cite{furlanello2018born} attempted to train a sequence of models, with the $i$-th model in the sequence being trained with knowledge distillation with the $i-1$-th model as the teacher.
They find that such sequential knowledge distillation may improve the performance compared to a model trained from scratch, and ensembling the sequence produces a better model. 

We first test this claim on CIFAR with multiple networks and with both knowledge distillation and attention transfer (Table~\ref{tab:sequential}).
We find that there are several caveats to Furlanello et al.'s result.
First, for some models (ResNet 8 and ResNet 14), the last student in the sequence actually underperforms a student model trained from scratch.
This suggests that the network architecture heavily determines the success of sequential knowledge distillation.
Second, we find that although an ensemble of the student models from the entire sequence outperforms a \emph{single} model trained from scratch, it does not outperform an ensemble of an equal number of models trained from scratch.
This might be because the students obtained through a sequence of knowledge distillation steps may be correlated with each other and therefore may not produce a strong ensemble.

\begin{table}
\begin{center}
\resizebox{0.9\linewidth}{!}{
\begin{tabular}{llccc}
\toprule
 {Method}&{Teacher}& Top-1 Error (\%) \\
\midrule
Scratch & - & 30.24   \\
Full KD~\cite{KD} & ResNet18 & 30.57 \\
Full KD~\cite{KD} & ResNet34 & 30.79 \\
Full KD~\cite{KD} & ResNet50 & 30.95  \\
Seq. Full KD~\cite{yang2018knowledge} & 3 Gen. & 30.12$^*$\\
Seq. Full KD~\cite{yang2018knowledge} & 6 Gen. & 29.6$^*$\\
KD+ONE~\cite{lan2018knowledge} & 3 Branches& 29.45 $\pm$0.23$^*$ \\
Full KD + AT~\cite{attentiontransfer}& ResNet34 & 30.94 \\
Full KD + AT~\cite{attentiontransfer}& ResNet34 & 29.3$^*$\\

\midrule
ESKD & ResNet18 & 29.01\\
ESKD & ResNet34 & 29.16\\
ESKD & ResNet50 & 29.35\\
ESKD & ResNet152 & 29.45\\
ESKD & ResNet34 (50) & 29.02\\
ESKD & ResNet50 (35) & 29.05\\
ESKD & ResNet152 (35) & 29.26\\
Seq. ESKD & L $\rightarrow$S$\rightarrow$S & 29.41 \\
Seq. ESKD & M$\rightarrow$S$\rightarrow$S & 29.35 \\
Seq. ESKD & S $\rightarrow$S$\rightarrow$S & 29.15 \\
\midrule
ESKD + AT & ResNet34 & 28.84\\
ESKD + AT & ResNet34 (50) & \textbf{28.61}\\
\bottomrule
\end{tabular}%
}
\end{center}
\caption{Overall result of ImageNet experiments. ESKD: Early-stopped knowledge distillation, AT: Attention transfer~\cite{attentiontransfer}. The number inside the parentheses is the total number of epochs if teacher training is early-stopped.*Numbers reported in paper.  }
\label{tab:imagenetES}
\end{table}

If sequential knowledge distillation does indeed improve the accuracy of a model, a natural question to ask is if the resulting model forms a better teacher.
To evaluate this, we conducted the following experiment.
We chose WRN16-1 as the student model and WRN16-3 as the teacher (note that this is the optimal teacher for this student as suggested by Figure~\ref{fig:TSoptimization}). 
We then trained the teacher using a sequence of 5 iterations of knowledge distillation. 
We compared the efficacy of this model as a teacher compared to a teacher trained from scratch.
As shown in Table~\ref{tab:sequentialteacher},
a teacher trained with a sequence of knowledge distillation iterations, though more accurate, is not in fact a better teacher.

As discussed above, we might be interested in a variant of this idea where we first attempt to distill from a ``large'' model to a ``medium'' model, and then from the ``medium'' model to a ``small'' model.
If this worked, it might help us avoid the issue of differing student and teacher capacities.
We compare this step-wise knowledge distillation to directly distilling from the large model to the small model, or from the medium model to the small model.
This might be a way to get around the effect we observed in Figure~\ref{fig:TSoptimization}, where the larger models were not necessarily better teachers.
We performed this experiment using WRN16-1 as the small model, WRN16-3 (the optimal teacher for WRN16-1) as the medium model and WRN16-8 as the large model.
We find that such a step-wise distillation does not work: it performs almost exactly the same as directly using the large model for distillation with the small model (Table~\ref{tab:largetosmall}).
Sequential distillation cannot help make large models better teachers.

We repeat some of these experiments on ImageNet and show the results in Table~\ref{tab:imagenetsequential}.
We use early-stopping when performing knowledge distillation based on results from the previous section.
Sequential knowledge distillation is in fact ineffective on ImageNet, too.
The best result corresponds to a single knowledge distillation from the ``small'' model to another ``small'' model, where ``small'' is ResNet18, ``Med.'' is ResNet50, and ``Large'' is ResNet152.
All these results suggest that despite the initial promise of sequential distillation, it is not a panacea and it especially does not help us use a large teacher to train a small student of significantly different capacity.

%Something wrong with the tables below.

\subsection{Early-stopped teachers make better teachers}

In the previous section we have shown that sequential knowledge distillation is ineffective. This might be because it doesn't address the core problem: the solution the large teacher has found is simply not in the solution space of the small student.
The only solution is to find a teacher whose discovered solution is in fact reachable by the student.

We may perform grid-search to find the optimal teacher network architecture, but that is too expensive. 
Instead, we propose to \emph{regularize} the teacher when training it.
In particular, we propose to stop the training of the large teacher early.
There is some evidence that a large network trained with only a few epochs behaves as a small network, while still encompassing a greater search space than small network ~\cite{caruana2001overfitting, mahsereci2017early}.
This method is extremely simple and cheap, since only a third to fourth of the total number of epochs are needed.
\begin{figure}
\centering
\includegraphics[width=0.9\linewidth]{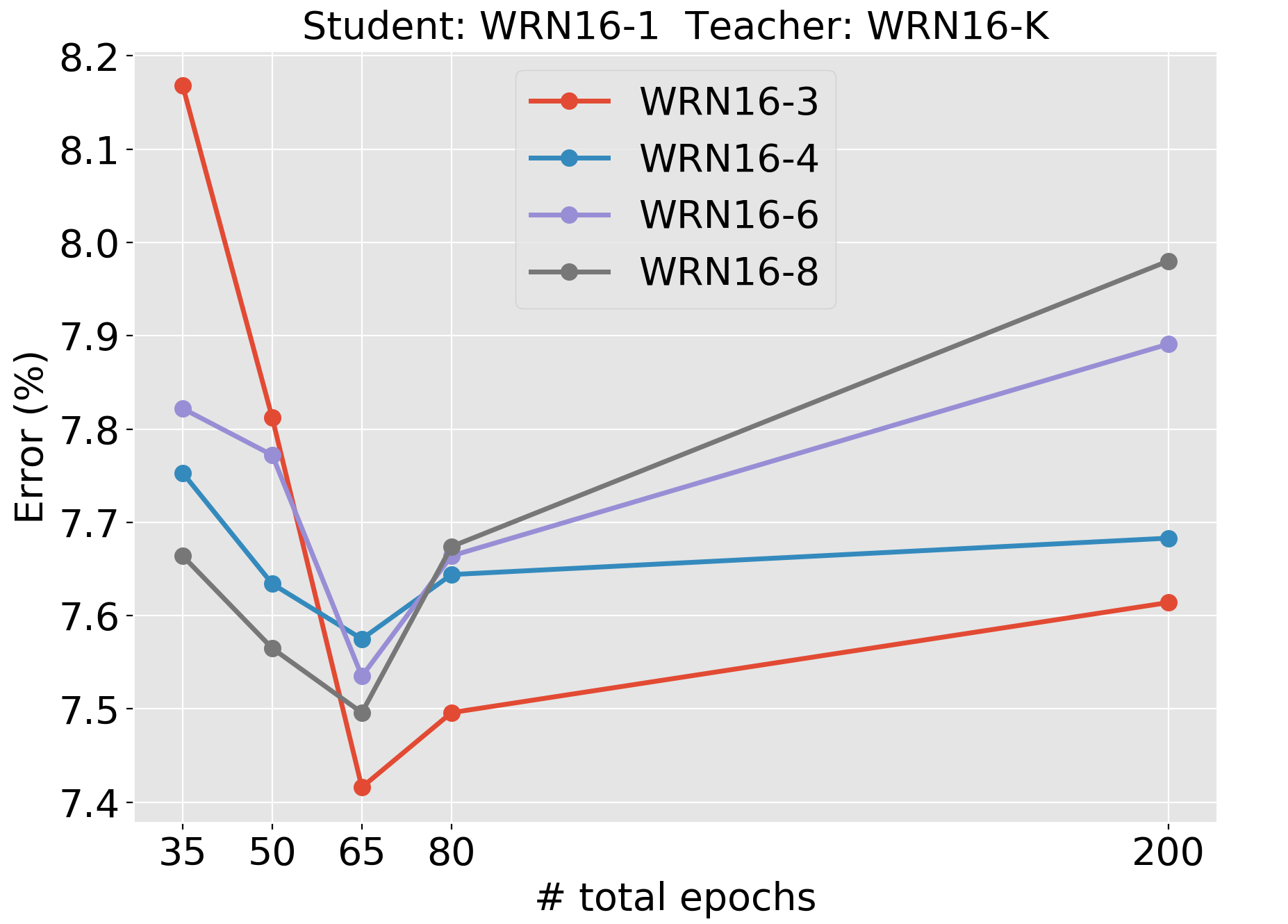}\\
\includegraphics[width=0.9\linewidth]{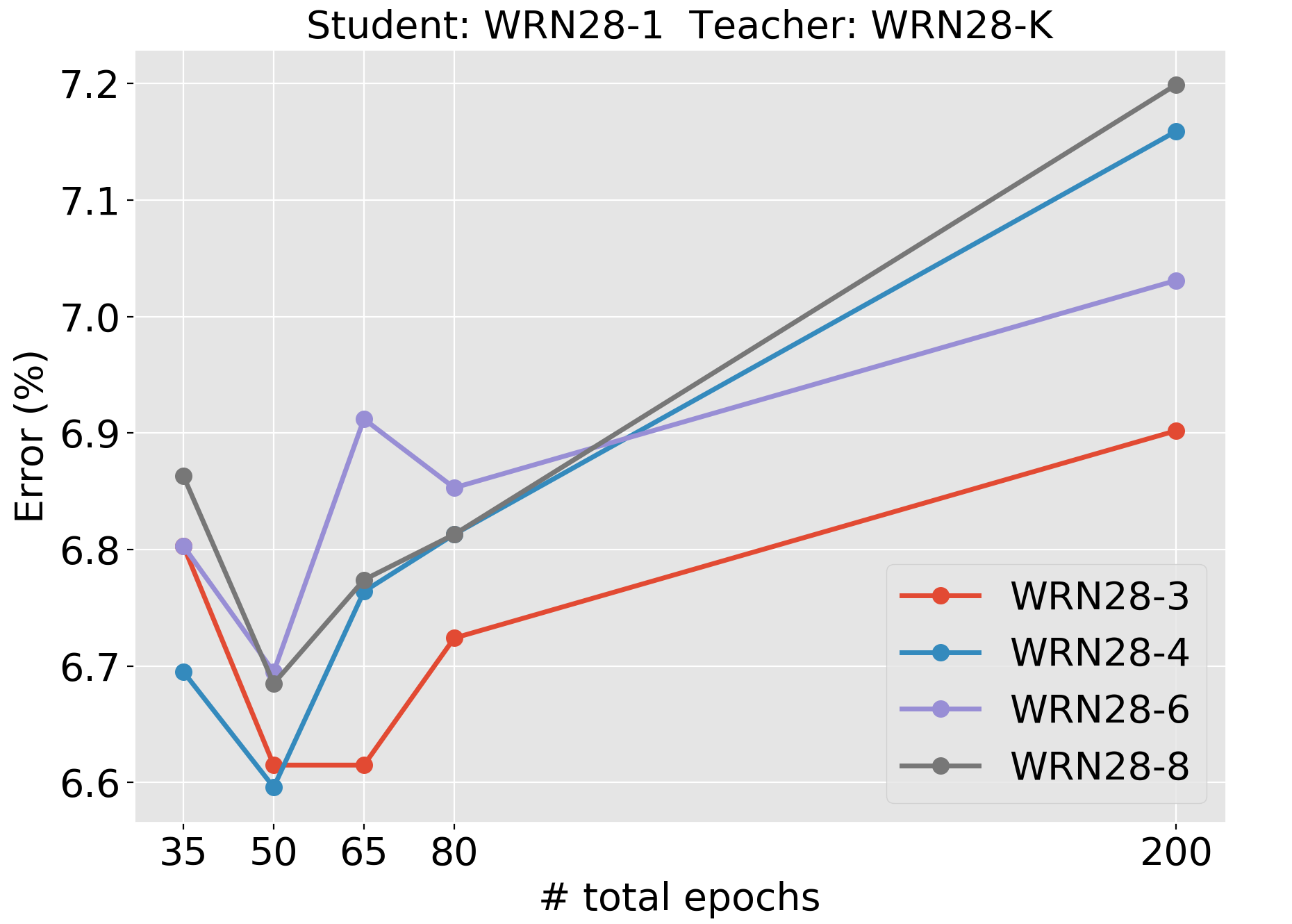}
\caption{ CIFAR10 result to examine the effectiveness of the knowledge distillation with early-stopped teachers. For both student types (WRN16-1 and WRN28-1), there are clear ``sweet spots'' which optimize the performance of student network.  }
\label{fig:early}
\end{figure}
\begin{figure*}
\begin{center}
\resizebox{\linewidth}{!}{
\includegraphics{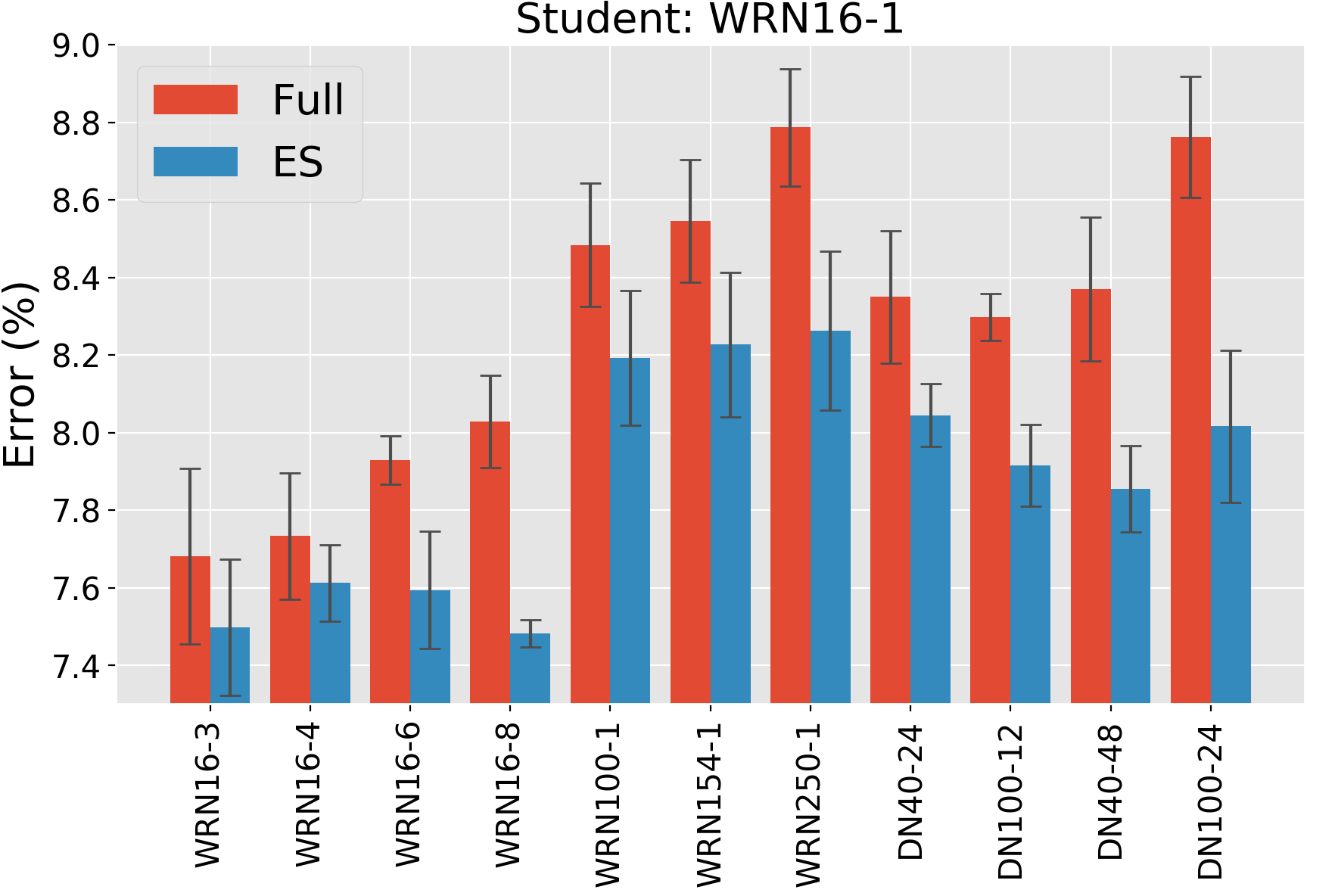}
\includegraphics{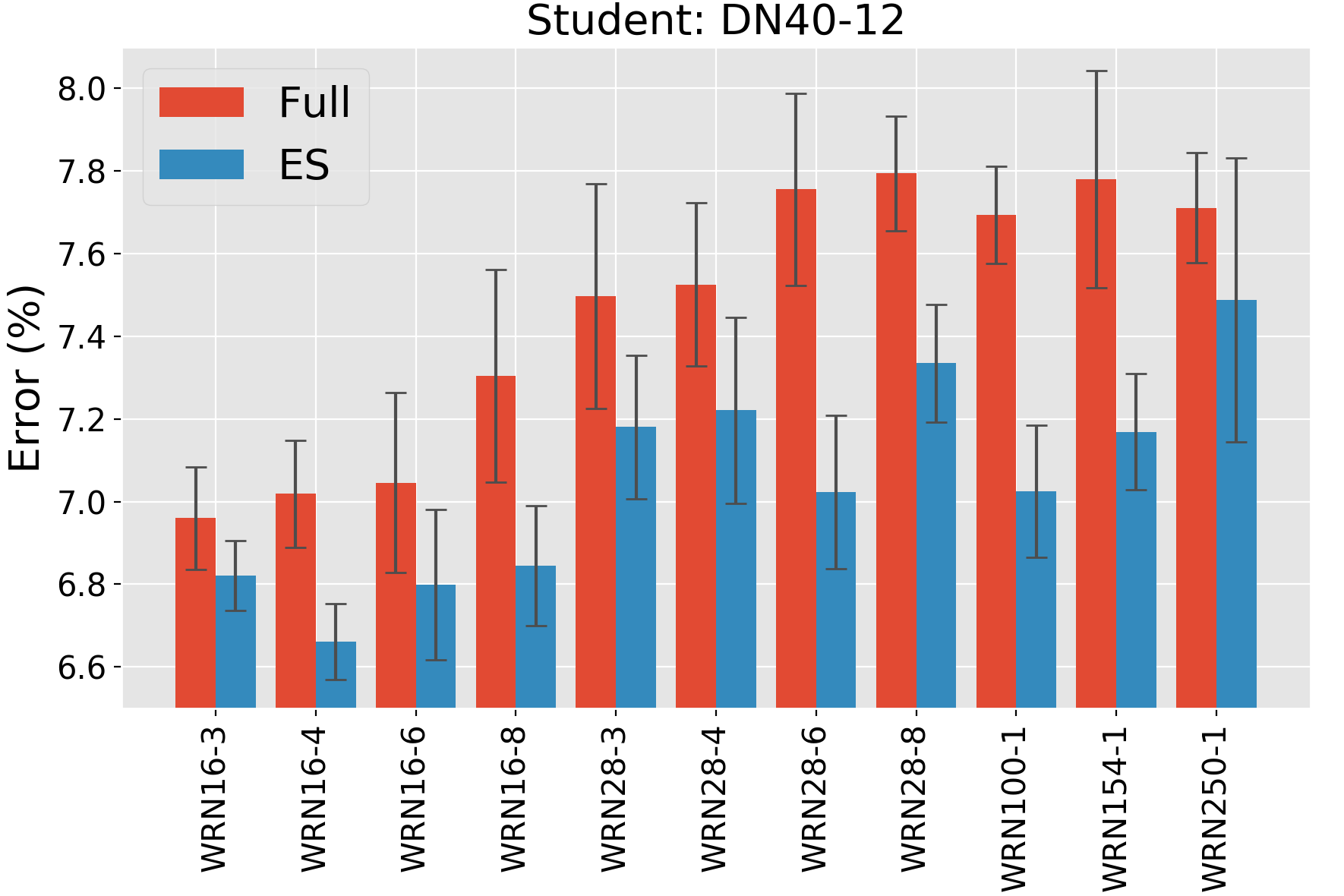}
}
\end{center}
\caption{In all different student and teacher configurations, students trained from early-stopped teacher consistently outperforms those trained from regular teacher. DenseNet40-12 and WideResNet16-1 were used as student network and DenseNet and WideResNet models with varying width and depth were used as teachers (x-axis)~\cite{densenet,wideresnet}. More results are in Supplementary. }
\label{fig:ablation2}
\end{figure*}
\begin{figure}
\begin{center}
\includegraphics[width=\linewidth]{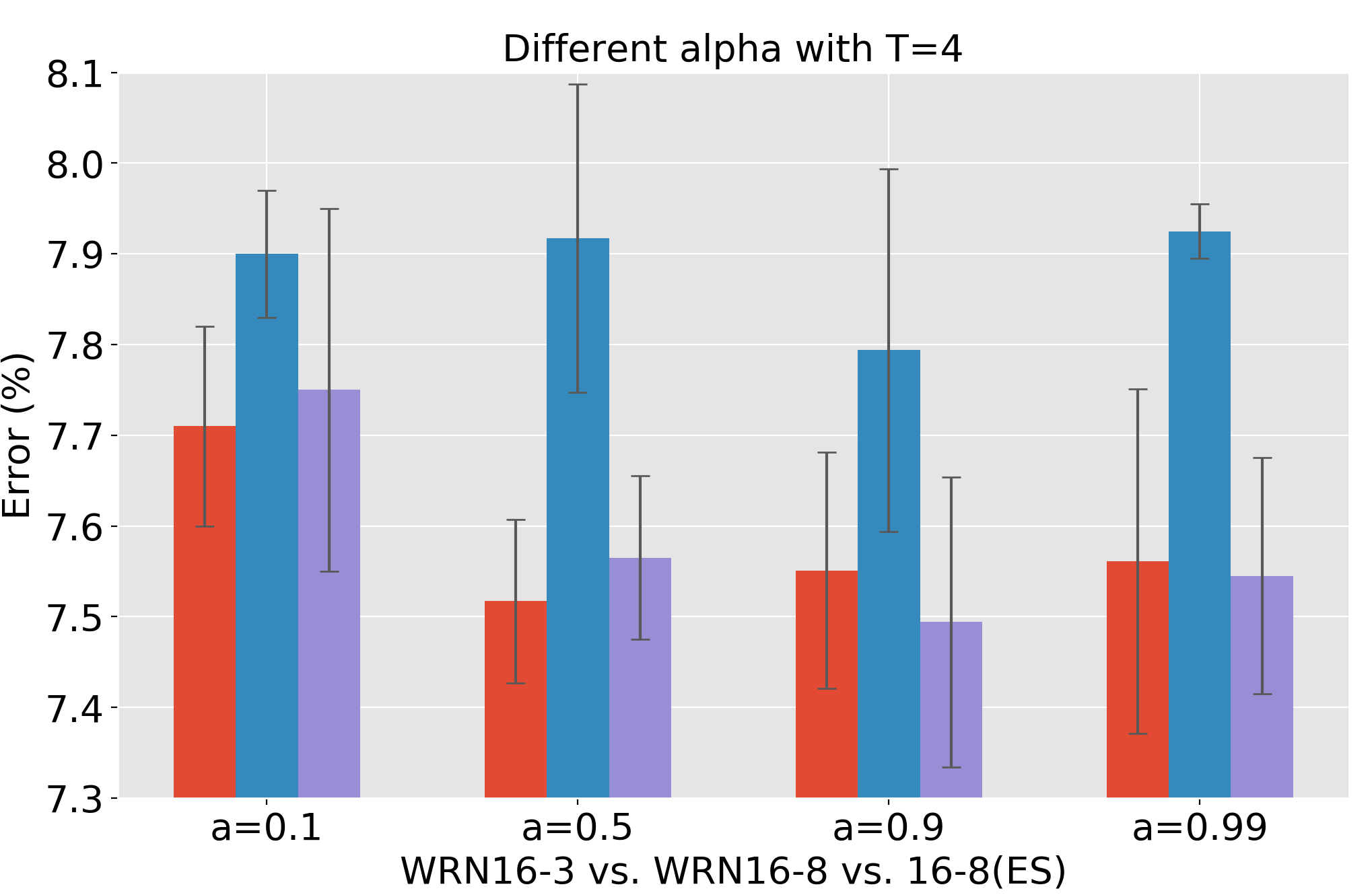}
\includegraphics[width=\linewidth]{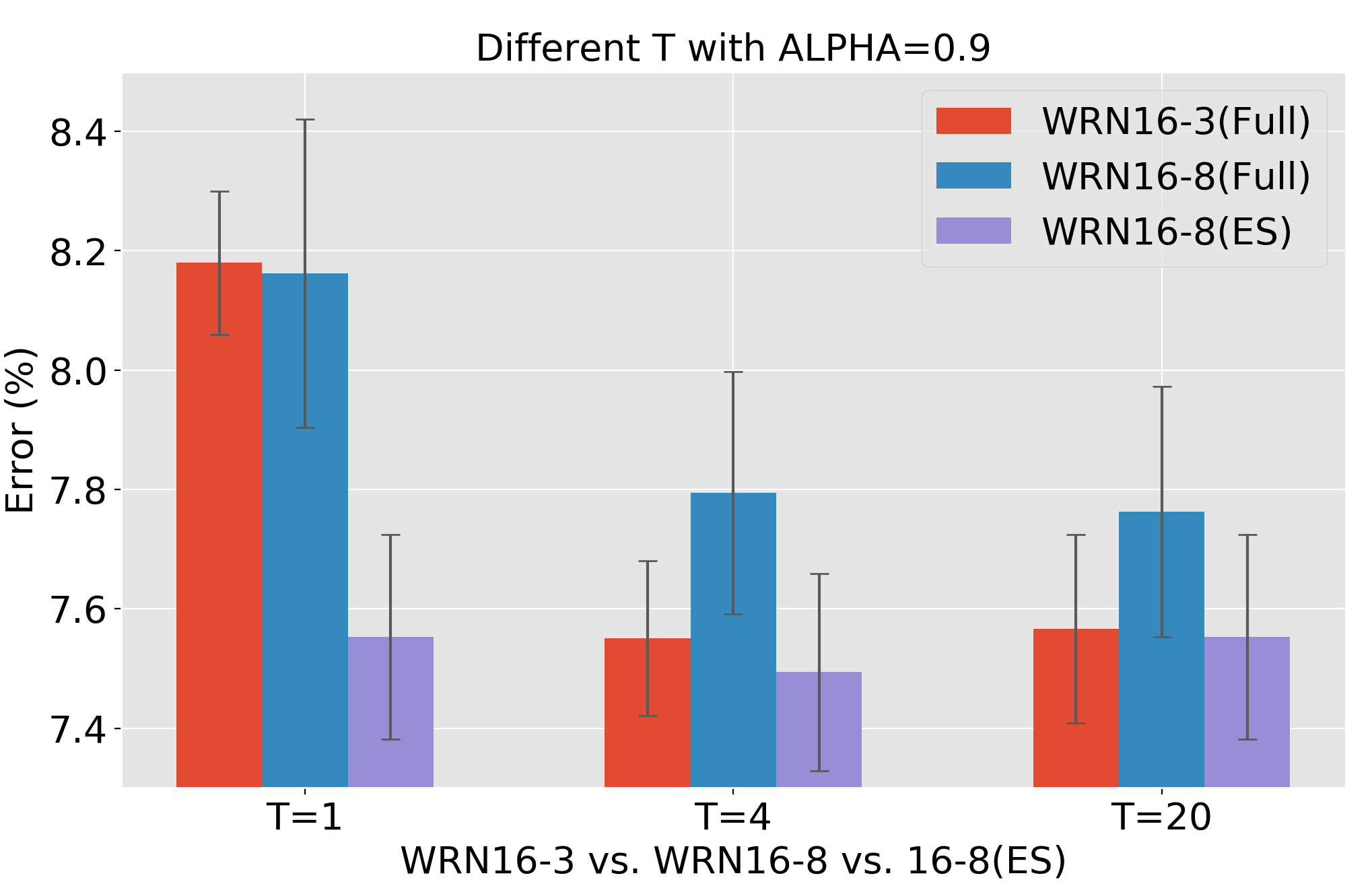}
\end{center}
\caption{ CIFAR10 result to examine that the effectiveness of using early-stopped network as a teacher is consistent to different hyperparameter settings. WRN16-8 (ES) has the same/better error compared to the optimal teacher WRN16-3. WRN16-1 is chosen as the student network.  }
\label{fig:ablation}
\end{figure}
We evaluate the effectiveness of this idea in both CIFAR10 and ImageNet. Figure ~\ref{fig:early} plots the error rates vs. total epochs on CIFAR10, where the x-axis represents the total number of epochs each teacher is trained. The same hyperparameters as other CIFAR10 experiments are used for the training teacher, except the total number of epochs and the learning rate schedule. For training the teacher network, the learning rate is dropped by $0.2$ every $\lfloor \frac{n-5}{3} \rfloor$ where $n$ is the total number of epochs. We chose $n\in \{35, 50, 65, 80\}$. Notice that for both student models (WRN16-1 and WRN28-1), all early-stopped teachers produce better students than the optimal fully-trained teacher (WRN16-3 and WRN28-3).  

Given these promising results, we next turn our attention to ImageNet. We choose $n\in \{35, 50\}$ and learning rate drop schedule of $(15, 25, 30)$ for 35 and $(20, 35, 45)$ for 50. Other hyperparameters and settings are the same with those of the previous ImageNet experiments.
Table~\ref{tab:imagenetES} shows results on ImageNet, where we also compare our results with prior results using knowledge distillation or its variants. Simply early-stopping the knowledge distillation with the largest, fully-trained teacher  outperforms most prior work $(\approx 29.45 \%)$. 
Our best teachers are the early-stopped ResNet34 and fully-trained ResNet18, ($\approx 29.01\%$) which has $\approx 1.23 $ point performance gain over the model trained from scratch and $\approx 0.2 \%$ from the best known result for this architecture from ~\cite{lan2018knowledge}. 

%We also find that the early-stopping ideas (both early-stopping of the teacher training, and early-stopping of the knowledge distillation) can be used together with other methods for improving knowledge distillation such as MMD-based intermediate feature mapping~\cite{attentiontransfer, NS, NIPS2017_6676}.
 Table~\ref{tab:imagenetES} also shows variants using attention transfer~\cite{attentiontransfer}, an improvement over knowledge distillation.
 Early stopping of the teacher and of the student are both very compatible with attention transfer, leading to improvements of 1.6 points over the baseline and 0.7 points over the best numbers obtained with attention transfer~\cite{attentiontransfer}.

\subsection{Other factors impacting knowledge distillation}
\paragraph{Different configurations}
In experiments above, we drew both the student and the teacher from the same model family.
We now experiment with teachers and students drawn from other, posssibly different, model families.
Figure~\ref{fig:ablation2} shows various combinations of DenseNets and Wide ResNets as students and teachers. Our conclusions, both the inefficacy of knowledge distillation from large teachers and the benefits from early stopping, are appaarent in these results.
%multiple student and teacher variants considering width, depth, and type of network as factors. Note that WideResNet with width factor equals to 1 reduces to Pre-activated ResNet~\cite{preresnet}. 

\begin{table}
\begin{center}
\resizebox{\linewidth}{!}{
\begin{tabular}{lccc}
\toprule
 Method & Teacher & Top-1 Error (\%) & Top-5 Error (\%)  \\
\midrule
Scratch & - & 47.38 & 18.51  \\
ESKD & ResNet18 & 47.09 & 18.13\\
Full KD & ResNet34& 47.86 & 18.61  \\
ESKD & ResNet34 (50) & 47.14 & 18.32  \\
Full KD & ResNet50& 47.92 & 18.72\\
ESKD & ResNet50 (35) & \textbf{47.02}& 18.14\\
ESKD & ResNet152 (35) & 47.25& 18.25\\
\bottomrule
\end{tabular}%
}
\end{center}
\caption{Each student network from ImageNet experiments is fine-tuned to Places-365 dataset for 12 epochs with initial learning rate $\gamma=0.1$ and weight decaying $10^{-1}$ every 3 epochs.}
\label{tab:places}
\end{table}

\begin{table}
\begin{center}
\resizebox{0.99\linewidth}{!}{
\begin{tabular}{lccc}
\toprule
Method & Teacher & Top-1 Error (\%) & Top-5 Error (\%)  \\
\midrule
Scratch & - & 30.24 & 10.92\\
Full KD & ResNet18 & 30.75 & 11.11\\
Full KD & ResNet50 & 30.98 & 10.20\\
Full KD & ResNet152 & 31.27 & 11.59 \\
\midrule
ESKD & ResNet18 & 29.00 & 9.91 \\
ESKD & ResNet50 & 29.00 & 9.76\\
ESKD& ResNet50 (35) & \textbf{28.89} & \textbf{9.76} \\
\bottomrule
\end{tabular}%
}
\end{center}
\caption{Experiments with temperature $\tau=20$ on IamgeNet dataset. High temperature increased overall results for ESKD methods (\textbf{lower half}) whereas had no difference for Full KD methods (\textbf{upper half}).}
\label{tab:highT}
\end{table}

\paragraph{Impact of $\alpha$ and $\tau$}
Till now we have fixed the tradeoff between KD and cross entropy, $\alpha=0.9$ and the temperature $\tau=4$. Although the standard choice of the temperature is $\tau \in \{3, 4, 5\}$, one might wonder if our conclusions about early stopping are sensitive to these choices.
As shown in Figure~\ref{fig:ablation} we find that the early-stopped teacher consistently outperforms the fully-trained teacher across a range of these hyperparameter values on CIFAR10. We further investigate the high temperature case on ImageNet dataset (Table~\ref{tab:highT}); we use $\tau=20$. High temperature can theoretically mitigate the peakiness of the teacher logits and may result better performance. We find that high temperature does increase the overall performance for early-stopped knowledge distillation (``ESKD'') but had no visible difference for full knowledge distillation (``Full KD''). The early stopped teacher still performed the best.

\paragraph{Generalizability for transfer learning}
Although we have seen variations in accuracies on ImageNet, a big aspect of convolutional networks is how well they transfer to other tasks.
In the table~\ref{tab:places} we examine whether the distilled network can be fine-tuned for classification on Places365 for a variety of students from the previous experiments. The results of transfer learning are consistent with the CIFAR and ImageNet experiments (full KD vs. early-stopped KD, small vs. large teachers, and regular vs. early-stopped teachers), proving that our findings also apply to transfer.

\section{Conclusion}
In this paper, we have presented an exhaustive study of the factors influencing knowledge distillation.
Our key finding is that knowledge distillation is not a panacea and cannot succeed when student capacity is too low to successfully mimic the teacher.
We have presented an approach to mitigate this issue by stopping teacher training early, to recover a solution more amenable for the student.
Finally we have shown the benefits of this approach on CIFAR10 and ImageNet and also on transfer learning on Places365.
We believe that further research into the nuances of distillation are necessary before it  can succeed as a general and practical approach.

\clearpage
{\small
\bibliographystyle{ieee_fullname}
\bibliography{main}
}

%%%%%%%%% TITLE
% \begin{titlepage}
\title{On the Efficacy of Knowledge Distillation - Supplementary Materials}

\author{Jang Hyun Cho\\
Cornell University\\
{\tt\small jc2926@cornell.edu}
\and
Bharath Hariharan\\
Cornell University\\
{\tt\small bh497@cornell.edu}
}
\maketitle
\ificcvfinal\thispagestyle{empty}\fi
% \end{titlepage}
%%%%%%%%% ABSTRACT

%%%%%%%%% BODY TEXT
\section{More Results on CIFAR10}
Here we report more results and details of experiments in our work. Consistent with the main paper, ``WRN'' and ``DN'' stand for WideResNet and DenseNet, respectively. Table~\ref{tab:wrn16} and \ref{tab:wrn28} show the efficacy of early-stopped teachers for student network WideResNet16-1 and WideResNet28-1 trained from teachers with varying width factor. As stated in the main paper, the number of total epochs $N\in \{35, 50, 65, 80, 200\}$ and learning rate decay step size $k \in \{10, 15, 20, 25, 60\}$ were considered in this experiment. Table~\ref{tab:cifar_full} shows that our conclusions are consistent with different knowledge distillation method such as attention transfer (``AT+KD''). 
% Table~\ref{tab:more_results} shows results with even shallower teacher networks (WRN16-4, WRN16-6, WRN16-11) chosen to be paired with WRN28-3, WRN28-4, and WRN28-8 in terms of the number of parameters. Table~\ref{tab:cosine} show more results with different settings.
Table~\ref{tab:wrnDN},~\ref{tab:DNDN},~\ref{tab:cosine}, and~\ref{tab:wrnDepth} show different experiment settings (different student-teacher pairs, learning method, etc.)

\section{Details on ImageNet Experiments}
\begin{figure}
\centering
\includegraphics[width=\linewidth]{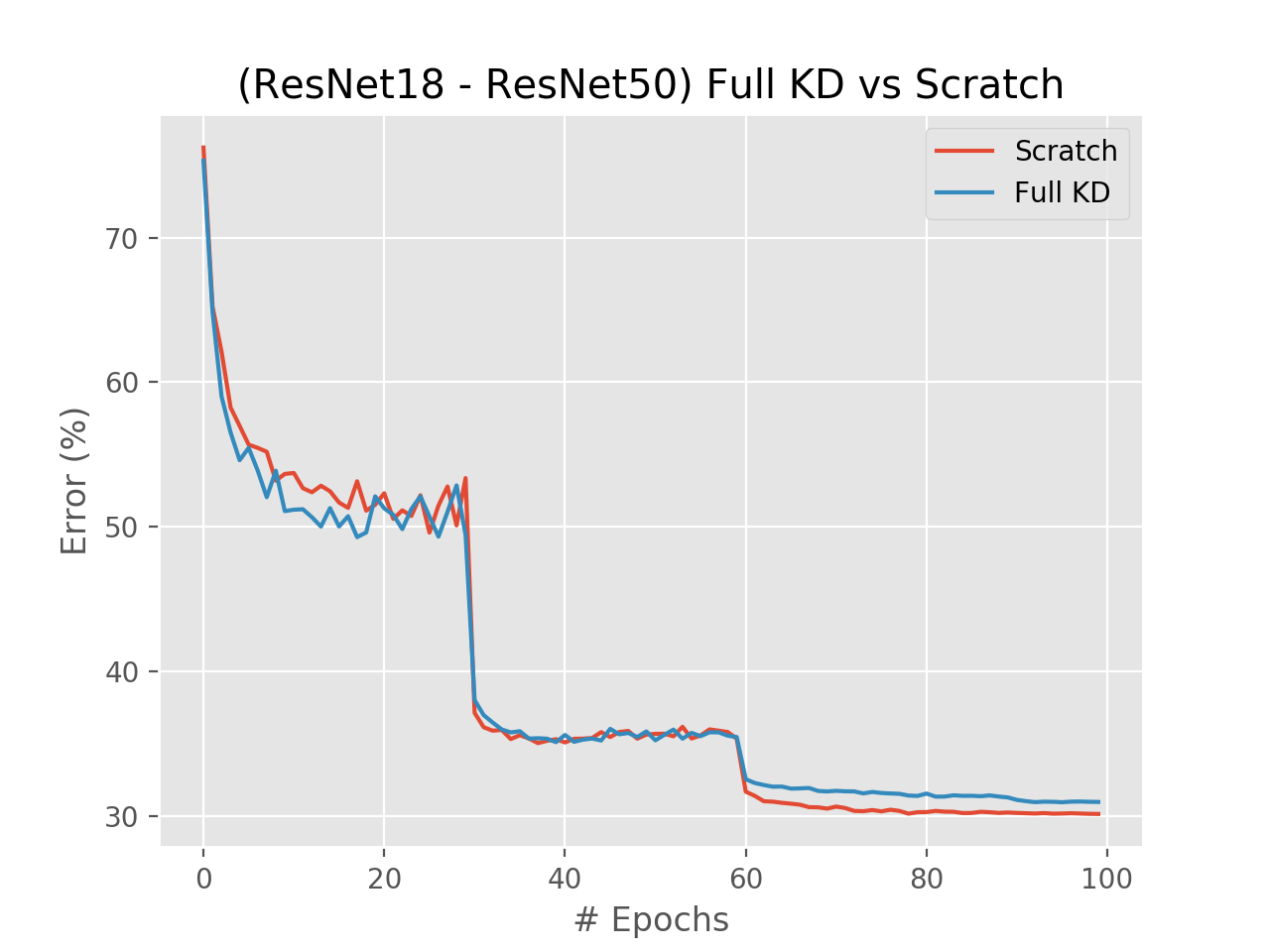}\\
\includegraphics[width=0.9\linewidth]{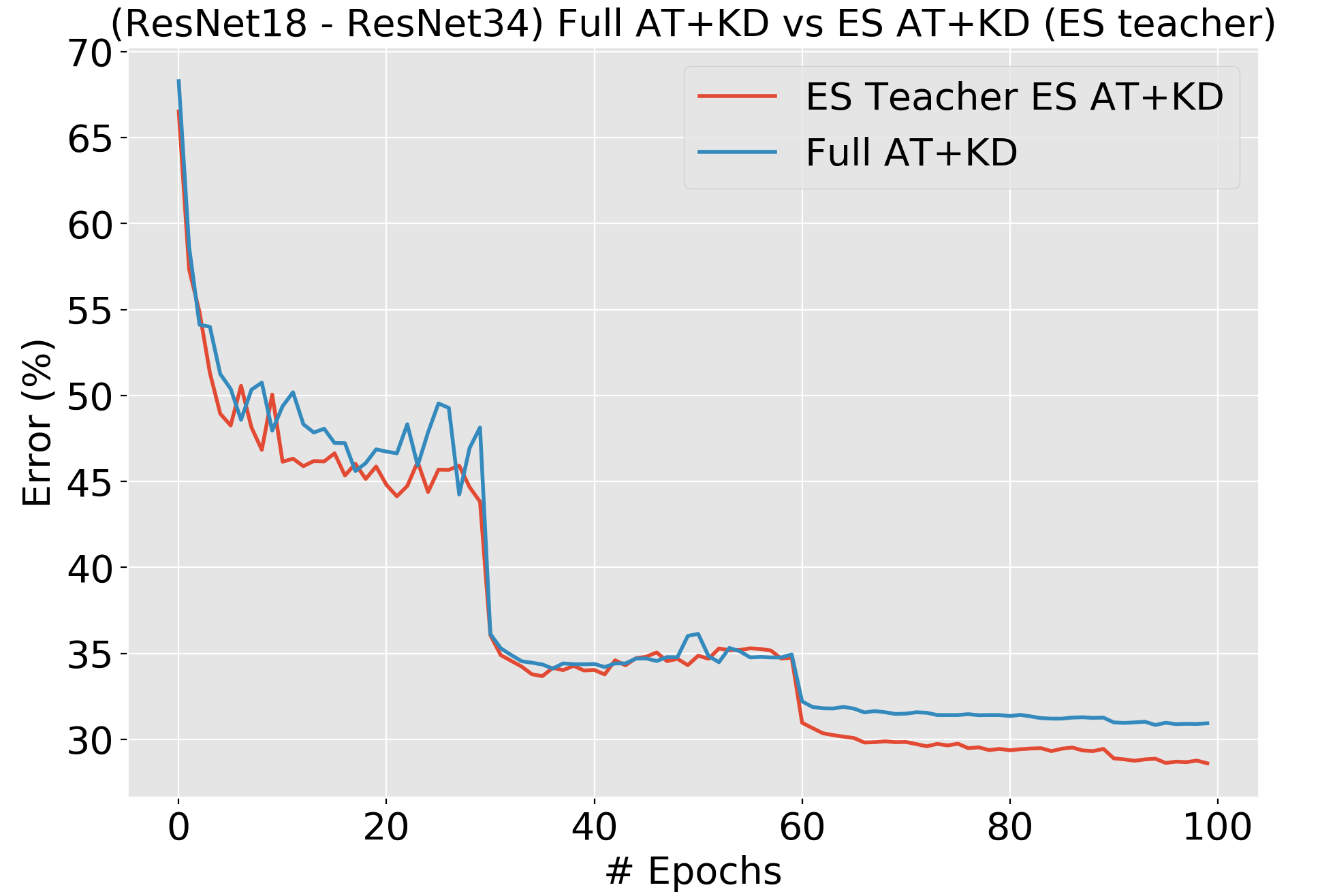}\\
\includegraphics[width=0.9\linewidth]{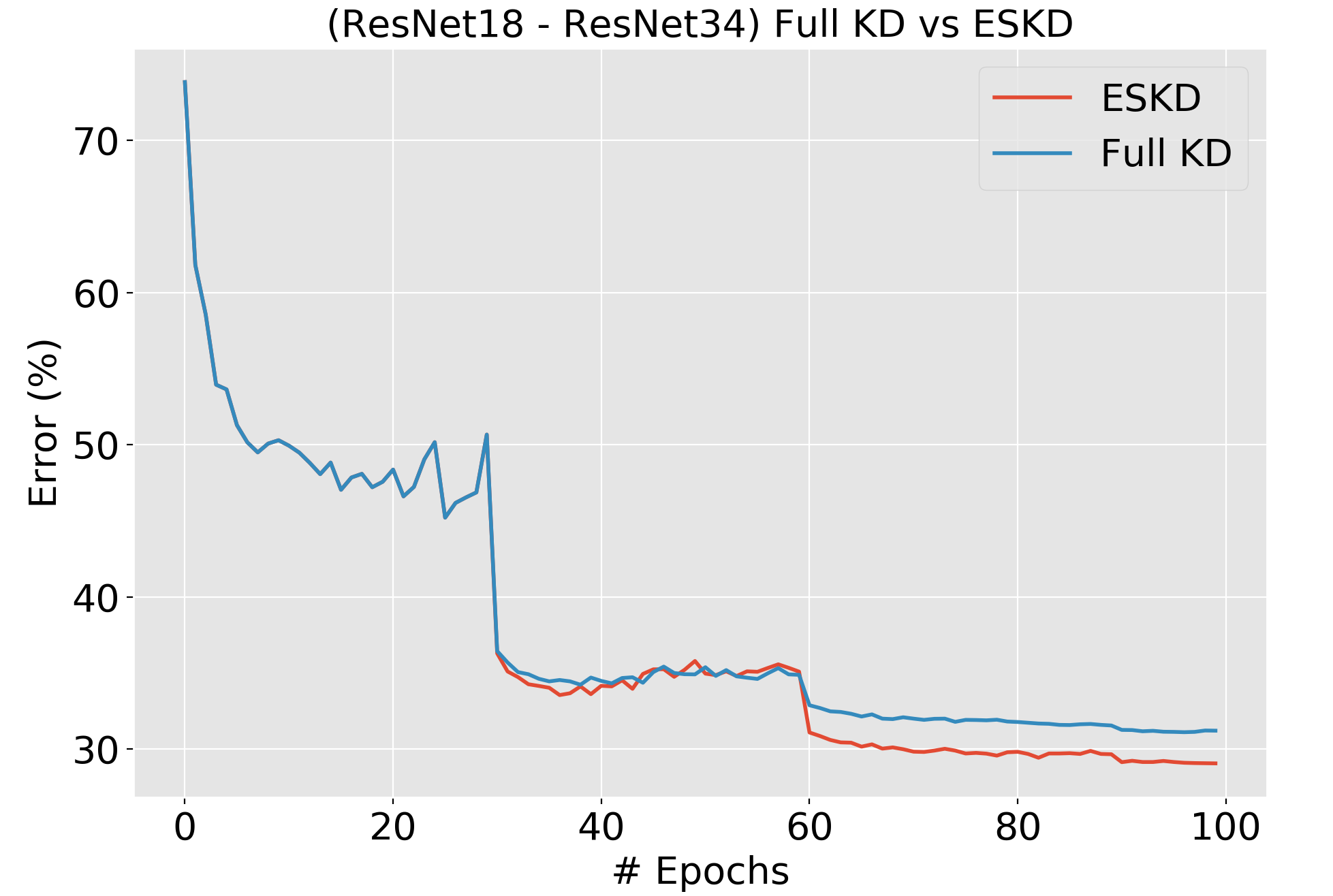}\\
\includegraphics[width=0.9\linewidth]{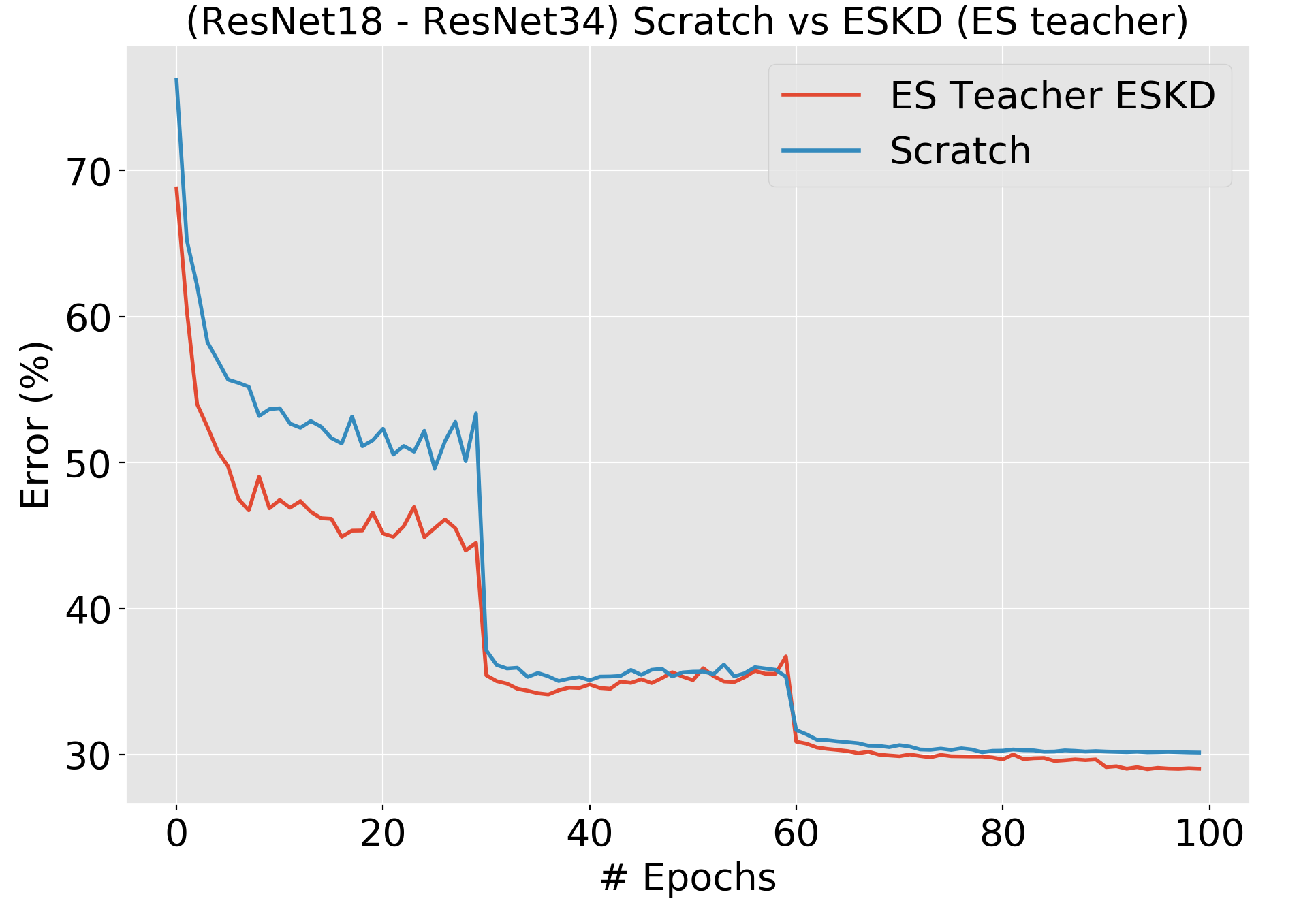}\\
\caption{ Different plots showing the harming effect of knowledge distillation when student capacity is limited, and how early-stopping mitigates the effect. }
\label{fig:plots}
\end{figure}

Here we report more details of ImageNet experiments. Figure~\ref{fig:plots} are comparisons of different student accuracy plots, showing the harming effect of distillation. Table~\ref{tab:teachers} shows the fully-trained and early-stopped models used as a teacher for ImageNet experiments. 

\begin{table}
\begin{center}
\resizebox{\linewidth}{!}{
\begin{tabular}{ccccccc}
\toprule
\multicolumn{5}{c}{\textbf{Student: WRN16-1}}\\
\toprule
Teacher &\# params & Mode& Teacher Error (\%) & Student Error (\%)\\
\midrule
- & - & 60/200 & - & 8.751 $\pm$ 0.129\\
\midrule 
DN40-24 & 0.69 M  & \multirow{4}{*}{60/200} & 5.419&  8.350 $\pm$ 0.195 \\ 
DN100-12 & 0.77 M & & 4.974 & 8.297 $\pm$ 0.069 \\
DN40-48 & 2.73 M& & 4.667 & 8.370 $\pm$ 0.212 \\ 
DN100-24 & 3.02 M & &4.272 & 8.763 $\pm$ 0.178 \\
\midrule
DN40-24 &0.69 M  &  \multirow{4}{*}{15/50}& 6.823 & 8.045 $\pm$ 0.092 \\ 
DN100-12 &0.77 M & & 6.615 & 7.915 $\pm$ 0.120 \\
DN40-48 &2.73 M & & 5.666 & \textbf{7.854} $\pm$ \textbf{0.127} \\ 
DN100-24 &3.02 M & & 5.435& 8.016 $\pm$ 0.223 \\
\bottomrule
\end{tabular}
}
\end{center}
\caption{WideResNet16-1 trained with different DenseNet teachers. First number next to ``DN'' indicates depth, followed by growth factor (consistent with the original paper). Top-row shows the result of WideResNet16-1 trained from scratch. In all cases, student trained with early-stopped DeseNet teacher performs better by large margin.  }
\label{tab:wrnDN}
\end{table}

\begin{table}
\begin{center}
\resizebox{\linewidth}{!}{
\begin{tabular}{ccccccc}
\toprule
\multicolumn{5}{c}{\textbf{Student: DN40-12}}\\
\toprule
Teacher &\# params & Mode& Teacher Error (\%) & Student Error (\%)\\
\midrule
- & - & 60/200 & - & 7.268 $\pm$ 0.148\\
DN40-12 & 0.18 M & 60/200 & 7.169 & 6.821 $\pm$ 0.226 \\
\midrule 
DN40-24 & 0.69 M  & \multirow{4}{*}{60/200} & 5.419&  6.964 $\pm$ 0.139 \\ 
DN100-12 & 0.77 M & & 4.974 & 6.847 $\pm$ 0.278 \\
DN40-48 & 2.73 M& & 4.667 & 7.266 $\pm$ 0.359 $^*$ \\ 
DN100-24 & 3.02 M & &4.272 & 7.507 $\pm$ 0.204 $^*$ \\
\midrule
DN40-24 &0.69 M  &  \multirow{4}{*}{15/50}& 6.823 & 6.981 $\pm$ 0.112 \\ 
DN100-12 &0.77 M & & 6.615 & \textbf{6.645} $\pm$ \textbf{0.089} \\
DN40-48 &2.73 M  & & 5.666 & 6.679 $\pm$ 0.123 \\ 
DN100-24 &3.02 M & & 5.435& 6.721 $\pm$ 0.298 \\
\bottomrule
\end{tabular}
}
\end{center}
\caption{DenseNet40-12 trained with different DenseNet teachers. First number next to ``DN'' indicates depth, followed by growth factor (consistent with the original paper). Top-row shows the result of DenseNet40-12 trained from scratch. In all cases student trained with early-stopped DeseNet teacher performs better by large margin. Numbers with $^*$ indicate that the students failed to achieve the same accuracy of student trained from scratch. }
\label{tab:DNDN}
\end{table}

\begin{table}
\begin{center}
\resizebox{\linewidth}{!}{
\begin{tabular}{cccc}
\toprule
{Student}& {Teacher} & Schedule Type & { Error (\%) }\\
\midrule 
\multirow{4}{*}{WRN16-1} & WRN16-8 & Cosine & 7.945 $\pm$ 0.127\\
& WRN16-8 (20/65) & Cosine & \textbf{7.781 }$\pm$ \textbf{0.201}\\
 & WRN100-1 & Cosine & 8.524 $\pm$ 0.182 \\
 & WRN100-1 (20/65) & Cosine & \textbf{8.191 }$\pm$ \textbf{0.104}\\
\bottomrule
\end{tabular}
}
\end{center}
\caption{CIFAR10 results of knowledge distillation with a different learning rate decaying schedule, ``Cosine'' scheduling. Student trained with early-stopped teacher performed better.  }
\label{tab:cosine}
\end{table}

\begin{table}
\begin{center}
\resizebox{\linewidth}{!}{
\begin{tabular}{lcccc}
\toprule
Model & \# params & Top 1 Error (\%) & Top 5 Error (\%)  \\
\midrule
ResNet18 & 11.69 M  & 30.24 & 10.92\\
ResNet34 &21.79 M  & 26.70 & 8.58 \\
ResNet34 (50) & 21.79 M & 27.72 & 9.10\\
ResNet50 & 25.56 M & 23.85 & 7.13 \\
ResNet50 (35) & 25.56 M&  27.01 & 8.75\\
ResNet152 & 60.19 M & 21.69 & 6.03\\
ResNet152 (35) & 60.19 M & 23.58 & 7.03 \\
\bottomrule
\end{tabular}
}
\end{center}
\caption{Details of models trained from scratch that are used as teachers for ImageNet experiments in the main paper. Models with a number inside parentheses are early-stopped.}
\label{tab:teachers}
\end{table}

\begin{table}
\begin{center}
\resizebox{\linewidth}{!}{
\begin{tabular}{ccccccc}
\toprule
\multicolumn{5}{c}{\textbf{Student: WRN16-1}}\\
\toprule
Teacher &\# params & Mode& Teacher Error (\%) & Student Error (\%)\\
\midrule
- & - & 60/200 & - & 8.751 $\pm$ 0.129\\
\midrule 
WRN40-1 & 0.56M  & \multirow{8}{*}{60/200} & 6.517&  8.324 $\pm$ 0.111 \\ 
WRN52-1 & 0.76 M & & 6.042 & 8.481 $\pm$ 0.198 \\
WRN64-1 &0.95 M & & 6.032 & 8.573 $\pm$ 0.158 \\ 
WRN76-1 &1.15 M  & &5.864 & 8.666 $\pm$ 0.121 \\
WRN88-1 &1.34 M & &5.686 & 8.811 $\pm$ 0.153\\
WRN100-1 &1.54 M & & 5.568 & 8.484 $\pm$ 0.182\\
WRN154-1 & 2.41 M & & 5.478 & 8.546 $\pm$ 0.181\\ 
WRN250-1 & 3.97 M& &5.271 & 8.787 $\pm$ 0.173\\
\midrule
WRN100-1 & 1.54 M& \multirow{3}{*}{15/50}& 7.526 & \textbf{8.192} $\pm$ \textbf{0.198}\\
WRN154-1 &2.41 M & & 7.318 & 8.227 $\pm$ 0.212\\ 
WRN250-1 & 3.97 M& & 6.893& 8.263 $\pm$ 0.234\\
\bottomrule
\end{tabular}
}
\end{center}
\caption{WideResNet16-1 trained with teachers varying depth factor. All students trained with early-stopped teacher performed better than any of students trained from fully-trained teacher by large margin. Among ones with fully-trained teachers, larger models did not make better student. All results are consistent with our conclusions. Note that WideResNet with width factor 1 is equivalent to Pre-Activated ResNet. }
\label{tab:wrnDepth}
\end{table}

\begin{table}
\begin{center}
\resizebox{\linewidth}{!}{
\begin{tabular}{ccccccc}
\toprule
\multicolumn{5}{c}{\textbf{Student: DN40-12}}\\
\toprule
Teacher &\# params & Mode& Teacher Error (\%) & Student Error (\%)\\
\midrule
- & - & 60/200 & - & 7.268 $\pm$ 0.148\\
\midrule 
WRN40-1 & 0.56M  & \multirow{8}{*}{60/200} & 6.517&  7.389 $\pm$ 0.244 \\ 
WRN52-1 & 0.76 M & & 6.042 & 7.640 $\pm$ 0.204 \\
WRN64-1 &0.95 M & & 6.032 & 7.600 $\pm$ 0.247 \\ 
WRN76-1 &1.15 M  & &5.864 & 7.407 $\pm$ 0.137 \\
WRN88-1 &1.34 M & &5.686 & 7.642 $\pm$ 0.131\\
WRN100-1 &1.54 M & & 5.568 & 7.693 $\pm$ 0.134\\
WRN154-1 & 2.41 M & & 5.478 & 7.780 $\pm$ 0.299\\ 
WRN250-1 & 3.97 M& &5.271 & 7.711 $\pm$ 0.152\\
\midrule
WRN100-1 & 1.54 M& \multirow{3}{*}{15/50}& 7.526 & \textbf{7.025} $\pm$ \textbf{0.182}\\
WRN154-1 &2.41 M & & 7.318 & 7.169 $\pm$ 0.161\\ 
WRN250-1 & 3.97 M& & 6.893& 7.488 $\pm$ 0.291\\
\bottomrule
\end{tabular}
}
\end{center}
\caption{DenseNet40-12 trained with WideResNet teachers varying depth factor. All students trained with early-stopped teacher performed better than any of students trained from fully-trained teacher by large margin. Among ones with fully-trained teachers, larger models did not make better student. All results are consistent with our conclusions. Note that WideResNet with width factor 1 is equivalent to Pre-Activated ResNet.}
\label{tab:wrnDepth}
\end{table}

\begin{table*}
\begin{center}
\resizebox{0.5\linewidth}{!}{
\begin{tabular}{ccccccc}
\toprule
\multicolumn{5}{c}{\textbf{Student: WRN16-1}}\\
\toprule
Teacher &\# params & Mode& Teacher Error (\%) & Student Error (\%)\\
\midrule 
WRN16-1 &0.17M & 60/200 &8.751&8.182 $\pm$ 0.250\\
WRN16-2 &0.69M& 60/200 & 6.269 & 7.610 $\pm$ 0.222\\
\midrule
\multirow{5}{*}{WRN16-3} & \multirow{5}{*}{1.55M}&60/200&5.340 & 7.681 $\pm$ 0.259\\
 & &25/80&6.289 & 7.517 $\pm$ 0.212\\
 & &20/65&6.507& \textbf{7.498} $\pm$ \textbf{0.201}\\
 & &15/50&6.734 & 7.788 $\pm$ 0.112\\
 & &10/35&7.416 & 8.093 $\pm$ 0.119\\
\midrule
\multirow{5}{*}{WRN16-4} & \multirow{5}{*}{2.74M}&60/200
         &4.964 & 7.733 $\pm$ 0.186\\
 & &25/80&5.666 & 7.658 $\pm$ 0.062\\
 & &20/65&5.963& \textbf{7.612} $\pm$ \textbf{0.112}\\
 & &15/50&6.358 & 7.788 $\pm$ 0.112\\
 & &10/35&7.130 & 8.093 $\pm$ 0.119\\
\midrule 
\multirow{5}{*}{WRN16-6} & \multirow{5}{*}{6.17M} &60/200 & 4.529 & 7.929 $\pm$ 0.071\\
&
&25/80  & 5.261 & 7.687 $\pm$ 0.157\\
&
&20/65  & 5.498 & \textbf{7.594} $\pm$ \textbf{0.173}\\
&
&15/50  & 5.893 & 7.685 $\pm$ 0.163 \\
&
&10/35  & 6.635 & 7.751 $\pm$ 0.157 \\
\midrule 
\multirow{5}{*}{WRN16-8} & \multirow{5}{*}{10.96M} & 60/200 & 4.410 & 8.028 $\pm$ 0.136\\
&
&25/80 & 4.984 & 7.642  $\pm$ 0.163 \\ 
& 
&20/65 & 5.270 & \textbf{7.482 }$\pm$ \textbf{0.041 }\\ 
& 
&15/50 & 5.498 & 7.596 $\pm$ 0.089 \\
&
&10/35 & 6.240 & 7.784 $\pm$ 0.223\\

\bottomrule
\end{tabular}
}
\end{center}
\caption{WideResNet16-1 trained with different teachers, and each teacher we performed different ``shrinking'' of the learning schedule. Step size $k\in\{10,15,20,25,60\}$ and total number of epoch $N\in \{35, 50, 65, 80, 200\}$ were considered. For kinds of teacher network, students trained with any of the early-stopped teachers outperforms the model trained with fully-trained teacher.}
\label{tab:wrn16}
\end{table*}

\begin{table*}
\begin{center}
\resizebox{0.5\linewidth}{!}{
\begin{tabular}{ccccccc}
\toprule
\multicolumn{5}{c}{\textbf{Student: WRN28-1}}\\
\toprule
Teacher &\# params & Mode& Teacher Error (\%) & Student Error (\%)\\
\midrule 
WRN28-1 &0.36M & 60/200 &7.101&7.101$\pm$ 0.072\\
WRN28-2 &1.46M& 60/200 & 5.201 & 6.973 $\pm$ 0.130\\
\midrule
\multirow{5}{*}{WRN28-3} & \multirow{5}{*}{3.29M}&60/200&4.687 & 6.952 $\pm$ 0.138\\
 & &25/80&5.369 & 6.702 $\pm$ 0.159\\
 & &20/65&5.696 & \textbf{6.621} $\pm$ \textbf{0.066}\\
 & &15/50&6.180 & 6.544 $\pm$ 0.262\\
 & &10/35&6.962 & 6.807 $\pm$ 0.076\\
\midrule
\multirow{5}{*}{WRN28-4} & \multirow{5}{*}{5.84M}&60/200
         &4.509 & 7.118 $\pm$ 0.198\\
 & &25/80&4.994 & 6.768 $\pm$ 0.099\\
 & &20/65&5.201& 6.772 $\pm$ 0.060\\
 & &15/50&5.824 & \textbf{6.610} $\pm$ \textbf{0.330}\\
 & &10/35&6.526 & 6.718 $\pm$ 0.063\\

\midrule 
\multirow{5}{*}{WRN28-6} & \multirow{5}{*}{13.14M} &60/200 & 4.104 & 7.070 $\pm$ 0.159\\
&
&25/80  & 4.608 & 6.869$\pm$ 0.152\\
&
&20/65  & 4.865 & 6.920 $\pm$ 0.114\\
&
&15/50  & 5.330 & 6.720 $\pm$ 0.060 \\
&
&10/35  & 5.992 & \textbf{6.710} $\pm$ \textbf{0.241}\\
\midrule 
\multirow{5}{*}{WRN28-8} & \multirow{5}{*}{23.25M} & 60/200 & 4.064 & 7.227 $\pm$ 0.149\\
&
&25/80 & 4.578 & 6.819  $\pm$ 0.155 \\ 
& 
&20/65 & 4.657 & 6.817$\pm$ 0.117\\ 
& 
&15/50 & 5.092 & \textbf{6.748 }$\pm$ \textbf{0.118} \\
&
&10/35 & 6.022 & 6.795 $\pm$ 0.123 \\

\bottomrule
\end{tabular}
}
\end{center}
\caption{WideResNet28-1 trained with different teachers, and each teacher we performed different ``shrinking'' of the learning schedule. Step size $k\in\{10,15,20,25,60\}$ and total number of epoch $N\in \{35, 50, 65, 80, 200\}$ were considered. For kinds of teacher network, students trained with any of the early-stopped teachers outperforms the model trained with fully-trained teacher.   }
\label{tab:wrn28}
\end{table*}

\begin{table*}
\begin{center}
\resizebox{0.9\linewidth}{!}{
\begin{tabular}{ccccccccc}
\toprule
{Student} &{\# params} &{Error (\%)}& {Teacher}  &{\# params} &{Teacher Error (\%)} & Method & { Student Error (\%) }\\
\midrule 
 & && WRN16-1 &0.17M&8.751 &KD & $8.182 \pm 0.250$ \\
 & & & WRN16-2 &0.69M&6.269& KD  & $\textbf{7.610} \pm \textbf{0.222}$ \\
\multirow{2}{*}{WRN16-1} &\multirow{2}{*}{0.17M}&\multirow{2}{*}{8.751}& WRN16-3 &1.55M&5.340 & KD & $7.681 \pm 0.259$ \\
&&& WRN16-4 &2.74M&4.964 & KD & $7.733 \pm 0.186$ \\
 & & & WRN16-6 &6.17M&4.529 &KD & $7.929 \pm 0.071$ \\
 & & & WRN16-8 &10.96M&4.410 & KD &$8.028 \pm 0.136$ \\
\midrule  
  & & & WRN16-2 &0.69M&6.269& AT+KD & $\textbf{7.498} \pm \textbf{0.062}$ \\
 & & & WRN16-3 &1.55M&5.340& AT+KD & $7.551 \pm 0.130$ \\
WRN16-1 &0.17M&8.751& WRN16-4 &2.74M&4.964& AT+KD & $7.656 \pm 0.131$ \\
 & & & WRN16-6 &6.17M&4.529& AT+KD & $7.668 \pm 0.139$ \\
  & & & WRN16-8 &10.96M&4.410& AT+KD & $7.794 \pm 0.203$ \\
\midrule
\multirow{4}{*}{WRN16-1} & \multirow{4}{*}{0.17M} & \multirow{4}{*}{8.751} & WRN16-3 (20/65) & 1.55M & 6.507 & AT+KD & 7.498 $\pm$ 0.201\\
 &  &                   & WRN16-4 (20/65) & 2.74M & 5.963 & AT+KD & 7.585 $\pm$ 0.165\\
 &  &                   & WRN16-6 (20/65) & 6.17M & 5.498 & AT+KD & \textbf{7.484 }$\pm$ \textbf{0.223}\\
 &  &                   & WRN16-8 (20/65) & 10.96M & 5.270 & AT+KD & 7.494 $\pm$ 0.165\\
\midrule 
 \multirow{6}{*}{WRN28-1}& \multirow{6}{*}{0.36M}& \multirow{6}{*}{7.101} &WRN28-1 &0.36M&7.101 & KD &$7.101 \pm 0.072$ \\
 & &  &WRN28-2 &1.46M&5.201&  KD &  $6.973 \pm 0.130$ \\
 &&   &WRN28-3 &3.29M&4.687 &KD & $\textbf{6.952} \pm \textbf{0.138}$ \\
&&&WRN28-4 &5.84M&4.509 & KD &$7.118 \pm 0.198$ \\
 & &  &WRN28-6 &13.14M&4.104 & KD &$7.070 \pm 0.159$\\
& & & WRN28-8 &23.35M&4.064 & KD & $7.227 \pm 0.149$ \\
\midrule 
\multirow{5}{*}{WRN28-1} &\multirow{5}{*}{0.36M} &\multirow{5}{*}{7.101}   &WRN28-2 &1.46M&5.201&  AT+KD &  $6.538 \pm 0.185$ \\
 &&  &WRN28-3 &3.29M&4.687 &AT+KD & $6.526 \pm 0.121$ \\
&&&WRN28-4 &5.84M&4.509 & AT+KD &$6.657 \pm 0.118$ \\
 & &  &WRN28-6 &13.14M&4.104 & AT+KD &$\textbf{6.443} \pm \textbf{0.092}$\\
& & & WRN28-8 &23.35M&4.064 & AT+KD & $6.487 \pm 0.222$ \\
\midrule
\multirow{4}{*}{WRN28-1} & \multirow{4}{*}{0.36M} & \multirow{4}{*}{7.101} & WRN28-3 (15/50) & 3.29M & 6.180 & AT+KD & 6.410 $\pm$ 0.162\\
 &  &                   & WRN28-4 (15/50) & 5.84M & 5.824 & AT+KD & 6.429 $\pm$ 0.090\\
 &  &                   & WRN28-6 (15/50) & 13.14M & 5.330 & AT+KD & \textbf{6.358} $\pm$ \textbf{0.168}\\
 &  &                   & WRN28-8 (15/50) & 23.35M & 5.092 & AT+KD & 6.402 $\pm$ 0.107\\
\bottomrule
\end{tabular}
}
\end{center}
\caption{WideResNet16-1 and WideResNet28-1 trained with teachers of increasing width. Attention transfer method was also explored. Teachers with ($k$/$N$) indicate early-stopped (step size/ total epochs). Our conclusions are consistent with different method such as attention transfer.  }
\label{tab:cifar_full}
\end{table*}

\begin{table*}
\begin{center}
\resizebox{0.9\linewidth}{!}{
\begin{tabular}{ccccccccc}
\toprule
{Student} &{\# params} &{Error (\%)}& {Teacher}  &{\# params} &{Teacher Error (\%)} & Method & { Student Error (\%) }\\
\midrule 
\multirow{4}{*}{WRN28-1} & \multirow{4}{*}{0.36M} & \multirow{4}{*}{7.101} & WRN16-4 &\multirow{4}{*}{2.74M}& 4.964 & KD & 6.518 $\pm$ 0.204 \\ 
 & & & WRN16-4 (20/65) & & 5.963 & KD & \textbf{6.483} $\pm$ \textbf{0.173}\\
 &  & & WRN16-4 && 4.964 & AT+KD & 6.357 $\pm$ 0.086 \\ 
  & & & WRN16-4 (20/65) &  & 5.963 & AT+KD &\textbf{ 6.253} $\pm$ \textbf{0.177}\\
\midrule 
\multirow{4}{*}{WRN28-1} & \multirow{4}{*}{0.36M} & \multirow{4}{*}{7.101} & WRN16-6 &\multirow{4}{*}{6.17M}& 4.529 & KD & 6.613 $\pm$ 0.227 \\ 
 & & & WRN16-6 (20/65) & & 5.498 & KD & \textbf{6.230} $\pm$ \textbf{0.069}\\
 &  & & WRN16-6 && 4.529 & AT+KD & 6.253 $\pm$ 0.278 \\ 
  & & & WRN16-6 (20/65) &  & 5.498 & AT+KD &\textbf{ 6.133} $\pm$ \textbf{0.113}\\
\midrule 
\multirow{4}{*}{WRN28-1} & \multirow{4}{*}{0.36M} & \multirow{4}{*}{7.101} & WRN16-11 &\multirow{4}{*}{20.70M}& 4.193 & KD & 6.774 $\pm$ 0.111 \\ 
 & & & WRN16-11 (20/65) & & 5.033 & KD & \textbf{6.281} $\pm$ \textbf{0.184}\\
 &  & & WRN16-11 && 4.193 & AT+KD & 6.360 $\pm$ 0.109 \\ 
  & & & WRN16-11 (20/65) &  & 5.033 & AT+KD &\textbf{ 6.202} $\pm$ \textbf{0.172}\\
\bottomrule
\end{tabular}
}
\end{center}
\caption{WideResNet28-1 student trained with even shallower teachers (WideResNet16-x) on CIFAR10. Consistent with our conclusions, early-stopped teachers produce better student. The teachers are chosen to be compared to WRN28-3, WRN28-4, and WRN28-8 in terms of the number of the parameters.}
\label{tab:more_results}
\end{table*}
\end{document}